%% file: main.tex
\documentclass[10pt]{article}

\usepackage[preprint]{tmlr}

\input{math_commands.tex}

\usepackage{hyperref}
\usepackage{url}
\usepackage{graphicx}
\usepackage{wrapfig}
\usepackage[table]{xcolor}
\usepackage{booktabs}
\usepackage{amsmath}
\usepackage{amssymb}
\usepackage{amsfonts}
\usepackage{subcaption}
\usepackage{caption}
\usepackage{makecell}
\usepackage{multirow}
\usepackage{tabularx}
\usepackage{enumitem}
\usepackage{adjustbox}
\usepackage{tcolorbox}
\usepackage{placeins}
\usepackage[percent]{overpic}
\hypersetup{
  hidelinks,
  pdftitle={MemeMind: Reference-Guided Trace Construction for Offline Context Optimization},
  pdfauthor={Run Yang, Weihang Wang, Boheng Sheng, Yuchen He, Jielei Zhang, Pengyu Chen, Zhiyu Wu, Qiang Sun, Huyang Sun, Longwen Gao}
}

\definecolor{enblue}{HTML}{003399}
\definecolor{sepgray}{gray}{0.6}
\definecolor{en_blue}{HTML}{003399}
\definecolor{sep_gray}{gray}{0.6}
\definecolor{stagepink}{HTML}{FFF4F5}
\definecolor{stageblue}{HTML}{E8F4FC}
\newcommand{\mres}[2]{#1{\color{sepgray}/}{\color{enblue}#2}}
\newcommand{\mresb}[2]{\textbf{#1}{\color{sepgray}/}\textbf{\color{enblue}#2}}
\newcommand{\method}{\textsc{MemeMind}}

\title{MemeMind: Reference-Guided Trace Construction for \mbox{Offline Context Optimization}}

\author{\name Run Yang\textsuperscript{1,*},
Weihang Wang\textsuperscript{1,2,*},
Boheng Sheng\textsuperscript{1,*}, Yuchen He\textsuperscript{1},
Jielei Zhang\textsuperscript{1,\textdagger} \\
\name Pengyu Chen\textsuperscript{1}, Zhiyu Wu\textsuperscript{2},
Qiang Sun\textsuperscript{1}, Huyang Sun\textsuperscript{1},
Longwen Gao\textsuperscript{1} \\
\addr \textsuperscript{1}Bilibili \quad
\textsuperscript{2}Fudan University \\
\addr \textsuperscript{*}Equal contribution.\quad
\textsuperscript{\textdagger}Corresponding author.
}

\begin{document}

\maketitle

\input{sections/0_abstract}
\input{sections/1_intro}
\input{sections/2_related_work}
\input{sections/3_methods}
\input{sections/4_datasets}
\input{sections/5_experiments}
\input{sections/6_conclusion}
\input{sections/7_impact}

\bibliography{references,references-added}
\bibliographystyle{tmlr}

\appendix
\input{appendix/appendix}

\end{document}

%% file: math_commands.tex
\usepackage{amsmath,amsfonts,bm}

\def\eqref#1{equation~\ref{#1}}

\def\1{\bm{1}}

\DeclareMathAlphabet{\mathsfit}{\encodingdefault}{\sfdefault}{m}{sl}
\SetMathAlphabet{\mathsfit}{bold}{\encodingdefault}{\sfdefault}{bx}{n}



%% file: sections/0_abstract.tex
\begin{abstract}
Offline context optimization improves an agent by revising its instructions and examples while keeping the model frozen.\footnote{Large language models were used to assist with language editing and typesetting. The authors are responsible for and verified all claims, analyses, and reported results.} This approach learns from rollouts on an adaptation set, but some queries produce only failed rollouts. In these cases, the optimizer sees no successful example of how the available tools can reach the correct answer. We introduce \method{}, which uses an offline reference answer to recover this missing experience. TraceBuilder identifies the evidence required by the reference, executes text search, image retrieval, and visual grounding, and verifies the resulting tool trace before adding it to the adaptation buffer. ToolGuide then summarizes the collected traces into a shared guide and separate instructions for each tool. The reference answers and constructed traces are used only during adaptation, while inference uses the learned guides with a frozen model. We study this problem through Anime, Comic, and Game meme interpretation. These memes combine edited and ambiguous visual content, overlaid text, long tail franchise knowledge, and culture specific references. Their interpretation can require coordinated visual grounding, image retrieval, and text search, making them a demanding setting in which native rollout groups may fail together. We evaluate \method{} on MemeX, a benchmark of 1,000 such memes annotated by experts. Across two Qwen3-VL models, two language partitions, and two independent judges, \method{} improves over the strongest context optimization baseline by 22.0\% and 21.1\% on Qwen3-VL-30B-A3B, and by 8.1\% and 8.0\% on Qwen3-VL-235B-A22B under GPT-5 judging. Ablations and held out traces show that constructing successful tool use for failed groups provides the largest component gain and produces more effective evidence acquisition at inference time.
\end{abstract}

%% file: sections/1_intro.tex
\section{Introduction}\label{sec:intro}

Tool augmented language and vision language agents can search the web, retrieve images, and inspect visual regions. Their performance often depends on instructions that specify when to use each tool and how to combine the returned evidence. Offline context optimization improves these instructions from an adaptation set while keeping the model parameters fixed~\citep{khattab2024dspy,agrawal2025gepa,zhang2025ace,cai2025tfgrpo}. It is useful when labeled data are limited, tools change frequently, or model training is impractical.

The quality of the learned instructions depends on the rollouts collected during adaptation. A rollout records the tool calls, observations, and final answer produced by an agent. Successful rollouts reveal procedures that can be reused on later queries, while failed rollouts can still provide comparisons or diagnostic feedback. However, when every sampled rollout for a query fails, the optimizer receives no example that shows how the original task can be solved. An outcome reference does not directly fill this gap. It states the correct interpretation but usually omits the regions to inspect, the queries to issue, and the evidence that connects tool results to the answer. Complete expert demonstrations provide these details~\citep{demoevolve}, but many datasets contain only reference answers. Table~\ref{tab:related-axes} isolates the capabilities needed to close this gap. Existing approaches provide only part of the chain. The problem is to turn a known answer into a verified tool use process, preserve the original task, and retain reusable tool instructions without exposing the answer when the adapted agent is deployed.

Anime, Comic, and Game meme interpretation provides a natural setting for studying this problem. A single meme may depend on character identity, source material, visual composition, translated text, and cultural references. An agent must often move between visual grounding, image retrieval, and text search before it can explain the meme. Ambiguous and long tail content can also cause a group of native rollouts to fail together. This setting tests whether offline adaptation can recover useful tool use from such groups rather than learning only from easier examples. More importantly, the task separates outcome knowledge from acquisition knowledge, since an expert explanation does not reveal which searches will recover its supporting evidence.

\begin{table*}[t]
\caption{Mechanism-level comparison. The fifth column asks whether a method produces an accepted process for the original goal from an all-failure group. Other failure signals may remain.}
\label{tab:related-axes}
\centering\scriptsize
\setlength{\tabcolsep}{3.5pt}
\resizebox{\textwidth}{!}{%
\begin{tabular}{@{}l l l c c c@{}}
\toprule
\textbf{Method} & \textbf{Optimized artifact} & \textbf{Process supervision} & \makecell{\textbf{Cross-query}\\\textbf{reuse}} & \makecell{\textbf{Original-goal process}\\\textbf{from all-failure group}} & \makecell{\textbf{Per-tool}\\\textbf{modularity}} \\
\midrule
ReAct & fixed prompt & none & $\times$ & $\times$ & $\times$ \\
ACE & context/playbook & execution feedback & $\checkmark$ & $\times$ & $\times$ \\
TF-GRPO & context prior & within-group preference & $\checkmark$ & $\times$ & $\times$ \\
DemoEvolve & external harness & complete demonstration & $\checkmark$ & $\times$ & $\times$ \\
HER / ECHO / AgentHER & memory or training data & relabeled alternative goal & $\checkmark$ & $\times$ & $\times$ \\
STaR & parametric policy & answer-conditioned rationale & $\checkmark$ & $\times$ & $\times$ \\
\rowcolor{gray!10}\textbf{MemeMind} & shared and tool-wise guides & outcome reference and execution & $\checkmark$ & $\checkmark$ & $\checkmark$ \\
\bottomrule
\end{tabular}}
\end{table*}

\begin{wrapfigure}[28]{r}{0.41\textwidth}
  \centering
  \vspace{-0.7\baselineskip}
  \includegraphics[width=\linewidth]{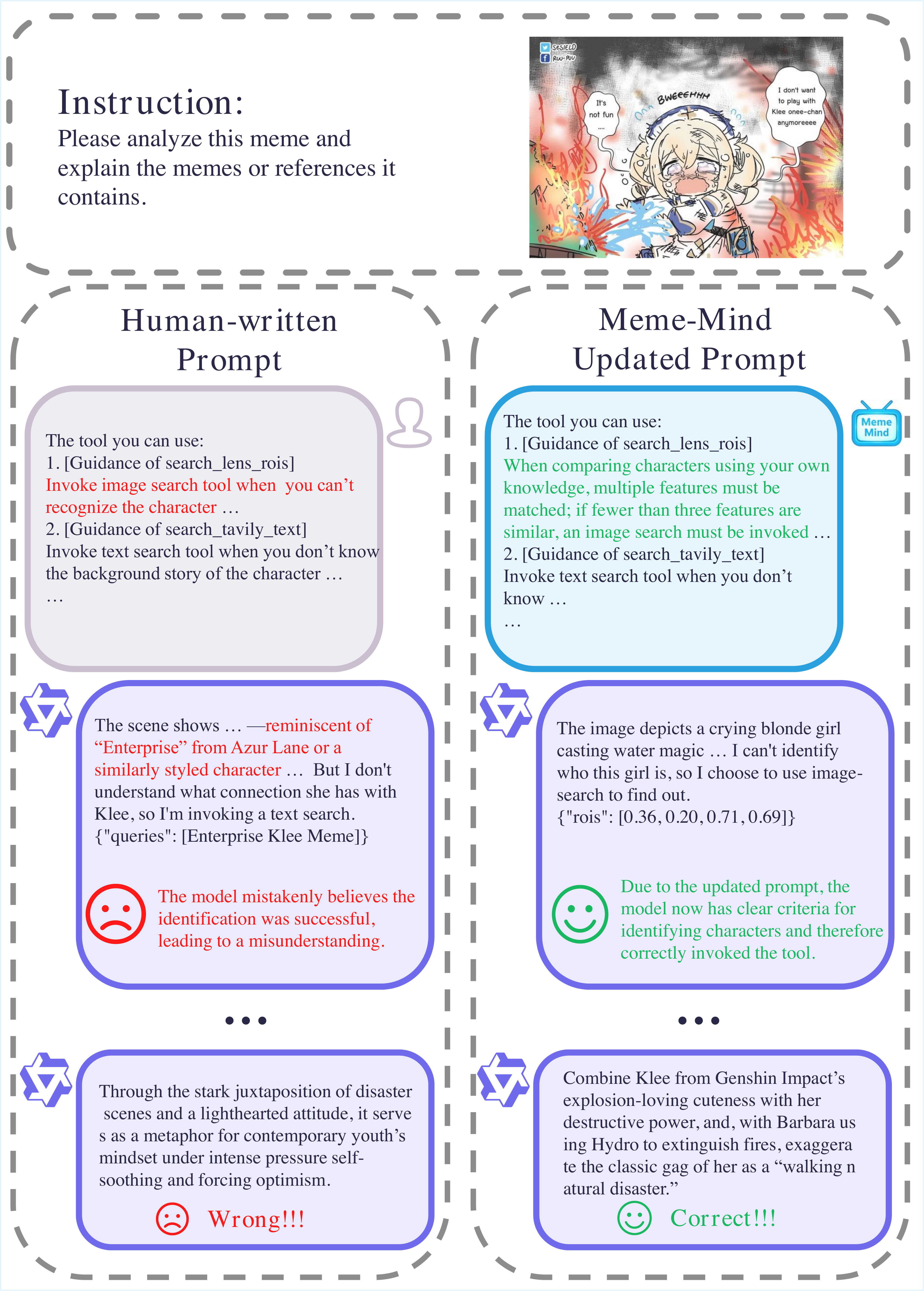}
  \caption{TraceBuilder constructs a successful trace after all native rollouts fail. ToolGuide converts collected traces into reusable shared and tool specific instructions.}
  \label{fig:meme}
  \vspace{-0.5\baselineskip}
\end{wrapfigure}

We introduce \method{} to recover the successful tool use that is missing from these failed groups. Figure~\ref{fig:meme} illustrates the two stages and shows how the learned instructions change the tool use of the agent. TraceBuilder uses the offline reference to identify target entities and regions. It plans and executes tool calls, links the returned observations to the reference, and admits the trace only after task and evidence verification. The constructed trace still addresses the original query, unlike hindsight methods that replace the original goal with one achieved by a failed trajectory~\citep{andrychowicz2017her,hu2025echo,agenther2026}. ToolGuide then converts the collected experience into instructions that can be reused without the references. A shared guide coordinates the overall interpretation, while separate guides specify when to use text search, image retrieval, and visual grounding, how to formulate their inputs, and how to pass evidence between them. These final guides are the only learned artifacts given to the frozen agent at inference time. This separation prevents the reference from becoming a deployment shortcut and forces the reusable artifact to encode acquisition decisions instead of case answers.

We evaluate \method{} on MemeX across two Qwen3-VL model scales, two language partitions, and two independent judges. Under GPT-5 judging, \method{} improves over the strongest context optimization baseline by 22.0\% and 21.1\% on Qwen3-VL-30B-A3B, and by 8.1\% and 8.0\% on Qwen3-VL-235B-A22B. Removing TraceBuilder lowers total score from 8.55 to 6.70 on the smaller model and from 9.32 to 8.80 on the larger model. These gains remain consistent across languages, judges, and model scales.

\WFclear
\newpage
\noindent\textbf{Contributions.} We identify the lack of a successful process when every rollout fails. We develop TraceBuilder and ToolGuide to recover this missing process and retain it as reusable tool instructions. We validate the approach across model scales, languages, judges, rollout budgets, and adaptation settings.

%% file: sections/2_related_work.tex
\section{Related Work}
\label{sec:related_work}

\subsection{Offline Context Optimization}

Context optimization steers model behavior through natural-language artifacts instead of parameter updates. DSPy treats prompts and demonstrations as optimizable components of a program~\citep{khattab2024dspy}. GEPA evolves prompts through reflective mutation and Pareto selection~\citep{agrawal2025gepa}, while ACE updates a playbook through generation, reflection, and curation~\citep{zhang2025ace}. Training-Free GRPO distills group-relative semantic advantages into a context prior for a frozen model~\citep{cai2025tfgrpo}. Dynamic Cheatsheet maintains compact external memory during interaction~\citep{suzgun2025dc}, and Meta Context Engineering jointly evolves context artifacts and the procedure that produces them~\citep{ye2026mce}. JTPRO further optimizes global tool instructions together with local tool descriptions~\citep{ghoshal2026jtpro}.

Together, these methods show that execution traces, scores, and reflective feedback can improve external context. They do not directly address process coverage when all rollouts for a query miss the acceptance criterion. Although relative or diagnostic feedback may remain, the group provides no accepted process trace for the original goal. \method{} targets this gap. TraceBuilder uses an outcome-level reference to construct the missing process and grounds it through the same tools available to native rollouts. ToolGuide then converts the collected experience into a global guide and tool-local guides, including explicit handoffs among visual grounding, image retrieval, and text search.

\subsection{Learning from Failed Experience}

Unsuccessful interaction can still provide learning signal through demonstrations, gold supervision, or hindsight. Complete demonstrations stabilize agentic harness evolution under sparse feedback~\citep{demoevolve}, and GCPO introduces gold supervision when contrastive rollout feedback is uninformative~\citep{wu2025gcpo}. Hindsight methods instead reinterpret a failed trajectory according to an alternative outcome that it achieved. HER relabels goals in reinforcement learning~\citep{andrychowicz2017her}, ECHO stores experience aligned with reached outcomes~\citep{hu2025echo}, and AgentHER constructs training data around achievable relabeled goals~\citep{agenther2026}. STaR provides the correct answer as a hint for rationale generation, then learns from the generated rationales through parameter updates~\citep{zelikman2022star}.

TraceBuilder follows a different supervision path. It keeps the original query fixed and uses the reference only to specify the desired outcome. It then plans a new evidence-acquisition process and executes heterogeneous tools to collect the required observations. A non-parametric guide optimizer consumes the accepted trace. The reference therefore moves the process toward the original goal, rather than moving the goal toward a failed process. This distinction matters for knowledge-intensive agents because a fluent answer-conditioned explanation does not show that the supporting evidence is obtainable through the available interfaces.

\subsection{Procedural Experience and Tool Modularity}

Reusable agent behavior can be externalized as examples, workflows, memories, or skills. ExpeL extracts cross-task insights and exemplars~\citep{zhao2024expel}, while Agent Workflow Memory induces reusable workflows from trajectories~\citep{wang2024awm}. Memp builds and updates procedural memory at multiple levels~\citep{fang2025memp}, and H-EPM organizes episodic experience around tool transitions~\citep{li2025hepm}. XSkill separates reusable skills from action-level experience for multimodal agents~\citep{jiang2026xskill}. These approaches establish procedural memory outside model weights as a practical form of agent adaptation.

ToolGuide organizes procedural reuse around the interfaces of a single agent. The shared guide records global interpretation principles, whereas each tool guide specifies invocation conditions, query construction, evidence tests, and output handoffs. This organization supports sequential multimodal search. Visual grounding produces regions and neutral descriptors, image retrieval tests identity hypotheses, text search resolves names, quotations, and source context, and the shared guide integrates the evidence into a final explanation. Because the learned artifact remains in context, test-time execution requires no additional retrieval policy.

\subsection{Knowledge-Intensive Multimodal and Meme Understanding}

Knowledge-aware visual QA benchmarks evaluate external knowledge, grounding, and relational reasoning~\citep{marino2019okvqa,hudson2019gqa,nayak2024benchmarking}. Meme benchmarks often focus on harmfulness and social meaning through classification~\citep{Kiela:2020hatefulmemes,pramanick2021momenta}. CHIME and MemeBridge address open-ended, cross-cultural explanation~\citep{xie-etal-2025-large,zhu2026memebridge}; MemeBench adds component-level diagnosis~\citep{wang2026memebench}.

MemeX isolates a retrieval-dependent form of meme understanding. Its ACG images combine edited panels, long-tail characters, localized phrases, and references drawn from games, animation, film, and online communities. Correct interpretation therefore depends on both entity acquisition and evidence synthesis. This setting can produce all-failure rollout groups and requires heterogeneous tool procedures. At the same time, expert explanations provide outcome supervision without revealing expert tool trajectories, making MemeX suitable for studying reference-guided process construction.

%% file: sections/3_methods.tex
\section{Method}\label{methods}

\begin{figure*}[t]
  \centering
  \begin{overpic}[width=0.96\textwidth]{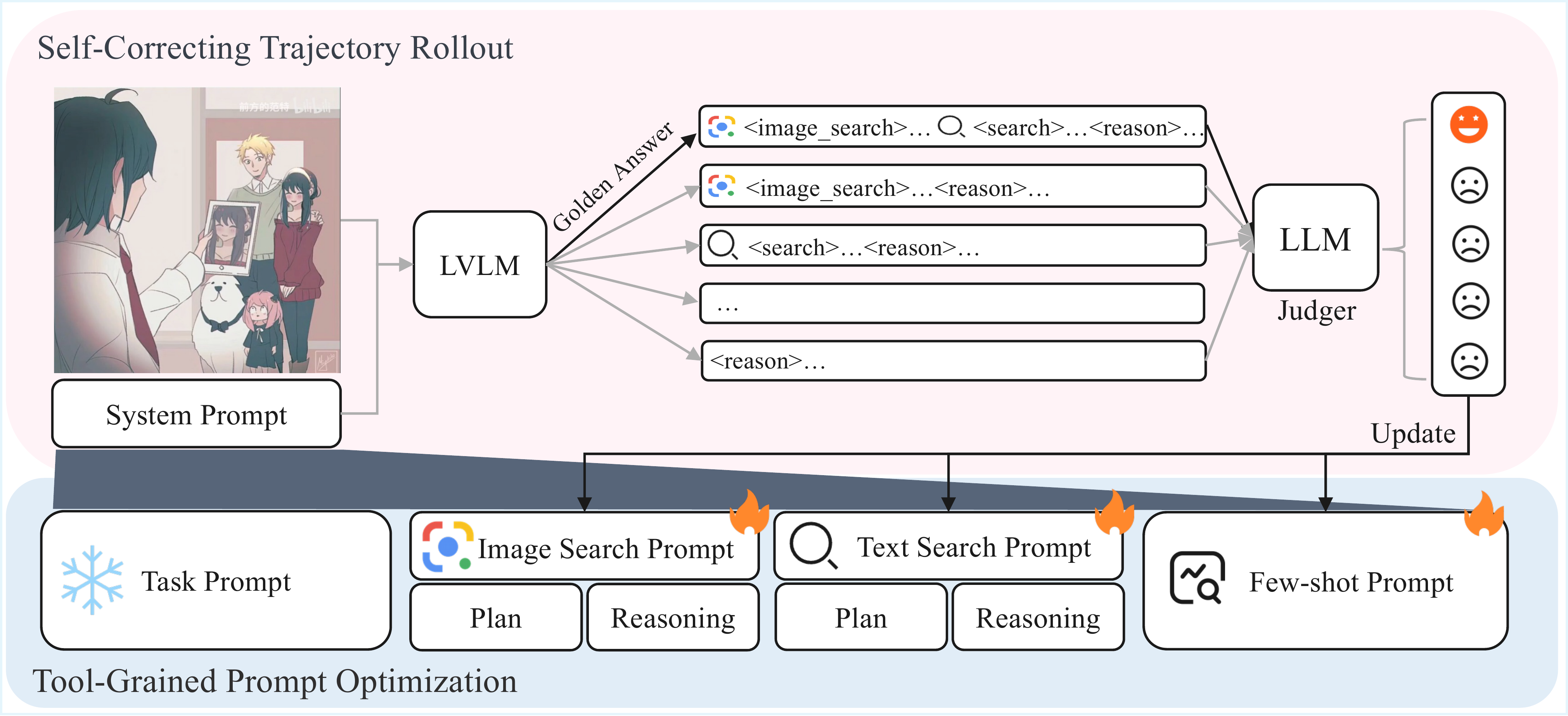}
    \put(1.8,42.4){\colorbox{stagepink}{\parbox{0.38\textwidth}{\centering\bfseries\large TraceBuilder}}}
    \put(1.8,1.5){\colorbox{stageblue}{\parbox{0.36\textwidth}{\centering\bfseries\large ToolGuide}}}
  \end{overpic}
  \caption{MemeMind has two stages. TraceBuilder constructs an accepted, tool-grounded trace for an all-failure group. ToolGuide learns a shared guide and per-tool guides from collected experience. No model parameters are updated.}
  \label{fig:framework}
\end{figure*}

Given a tool-augmented LVLM $\mathcal M$, judge $\mathcal J$, curator $\mathcal U$, and adaptation set $\mathcal D=\{(q_i,a_i)\}_{i=1}^N$, MemeMind optimizes natural-language guides $P$ while keeping model parameters fixed:
\begin{equation}
P^*=\arg\max_P\;\mathbb E_{(q,a)\sim\mathcal D}\!\left[\mathrm{Score}\bigl(\mathcal J(\mathcal M(q;P),a)\bigr)\right].
\end{equation}
The reference $a$ is available only during offline adaptation. MemeMind alternates tool-grounded trace construction and tool-wise guide learning. At test time, the model receives only the final $P$.

\paragraph{Design objective.}
MemeMind uses accepted process coverage as the collection objective. Let $A_{\mathcal J}(\tau;q,a)\in\{0,1\}$ indicate whether judge $\mathcal J$ accepts trace $\tau$ for the original query and reference pair $(q,a)$. Let $E(\tau)=1$ require every cited observation to come from an executed tool call. The full-coverage target for buffer $\mathcal B$ over $\mathcal D$ is
\[
\forall(q_i,a_i)\in\mathcal D,\quad
\exists\tau\in\mathcal B_i:\;
A_{\mathcal J}(\tau;q_i,a_i)\,E(\tau)=1 .
\]
An accepted native rollout satisfies this condition directly. When every rollout fails, TraceBuilder attempts to construct a trace for the same goal, providing a positive, tool-grounded process example where native exploration supplies none. Only traces that pass task-level acceptance and claim-level evidence checks enter the buffer.

\subsection{TraceBuilder: Reference-Guided Tool-Trace Construction}
For each query $q$, $\mathcal M$ samples $G$ multi-turn trajectories that interleave thoughts, tool calls, observations, and updated thoughts. The judge compares each final explanation with $a$ on character recognition, story understanding, and meme comprehension. Accepted trajectories and judge feedback enter the experience buffer.

When no trajectory clears the acceptance bar, TraceBuilder turns the known outcome into a grounded evidence-acquisition process. Given $(q,a)$, the model identifies characters, story elements, and regions of interest, formulates representative queries, and invokes the available tools in a coherent sequence. It assembles the returned observations into a trace explaining how each one supports $a$. The same three-dimension task judge used for native rollouts checks the interpretation. A claim-level verifier then checks whether each asserted identity, source, and contextual relation is supported by an executed tool observation. Unsupported claims trigger targeted queries and re-synthesis within the same five-round interaction limit used by native execution. Repair terminates when both checks pass or that limit is reached. The resulting accepted traces are grounded in actual tool observations rather than an answer-conditioned rationale alone.

The reference specifies \emph{what} the interpretation should contain, while TraceBuilder determines \emph{how} to acquire its evidence. The trace records reusable decisions, including region selection, entity hypotheses, query formulation, tool ordering, and evidence-to-claim links. TraceBuilder activates only when a group contains no accepted trajectory; otherwise, the buffer keeps accepted native executions.

\begin{table}[t]
\caption{TraceBuilder augments collection only at Step 3; evidence verification and bounded repair gate insertion of the constructed trace.}
\label{tab:tracebuilder-steps}
\centering\small
\begin{tabularx}{\columnwidth}{@{}cX@{}}
\toprule
\textbf{Step} & \textbf{Trace collection for query $(q,a)$} \\
\midrule
1 & Sample $G$ tool-augmented trajectories under $P^{(k)}$. \\
2 & Judge each trajectory against $a$ and retain accepted traces. \\
3 & If none is accepted, extract targets from $a$, plan and execute tool calls, and assemble the observed evidence into $\tau^+$. \\
4 & Verify task correctness and claim-level evidence, repair missing support within the five-round limit, and store $\tau^+$ only if both checks pass. \\
\bottomrule
\end{tabularx}
\end{table}

\subsection{ToolGuide: Tool-Wise Guide Learning}
ToolGuide represents $P$ as a shared global guide $P_g$ and per-tool guides $\{P_t\}$, each implemented as a natural-language protocol. The global guide provides cross-modal interpretation principles and examples. Each $P_t$ contains (i) a planning instruction for invocation and query construction and (ii) a reasoning instruction for evidence interpretation and integration. After every rollout iteration, $\mathcal U$ reads trajectories and judge feedback, then edits $P_g$ and $\{P_t\}$:
\[
P^{(k+1)}=\operatorname{Update}\!\left(P^{(k)};\mathcal B^{(k)},\mathcal J,\mathcal U\right).
\]
For example, the learned image-search protocol requires at least three discriminative features verified by two sources before it asserts an identity. Descriptors that remain unresolved are passed to text search as tentative hypotheses. The appendix provides the complete learned protocols and prompts.

\paragraph{Guide anatomy.}
The learned image guide first searches the full image and its salient regions, then records neutral descriptors. It withholds names unless facial or sprite similarity is corroborated by non-facial attributes and independent sources. The text guide treats OCR names as tentative. It confirms a template or scene through image retrieval before combining canonical template terms, character attributes, and exact quoted text in targeted queries. Image retrieval remains identity-neutral until sufficient evidence accumulates, while text retrieval consumes and verifies the resulting visual hypotheses. The shared guide integrates both procedures into a coherent final explanation.

\paragraph{Optimization and inference.}
During each offline epoch, the current guides are applied to the adaptation set and the resulting trajectories are judged. TraceBuilder supplies a trace when required, after which the curator makes constrained edits to the relevant guide components. GPT-5 is the default curator, while the sensitivity study varies this choice. Before selection, candidate guides are screened for copied entities, answer strings, and case-specific facts, then compared with the previous guides on a disjoint held-out validation split. The buffer and references are discarded once adaptation is complete. Inference augments the frozen backbone only with the final natural-language guides and ordinary tool calls.

%% file: sections/4_datasets.tex
\section{MemeX and Experimental Setup}\label{datasets}

\begin{wraptable}[17]{l}{0.47\textwidth}
  \centering
  \vspace{-0.4\baselineskip}
  \caption{MemeX compared with representative meme benchmarks.}
  \label{tab:dataset-comparison}
  \vspace{0.25em}
  \scriptsize
  \setlength{\tabcolsep}{3pt}
  \renewcommand{\arraystretch}{1.18}
  \resizebox{\linewidth}{!}{%
  \begin{tabular}{@{}lccc@{}}
  \toprule
  \textbf{Dataset} & \textbf{Modality} & \textbf{Task} & \textbf{Output} \\
  \midrule
  HarMeme & Image & toxicity & classification \\
  Harm-C/P & Image & toxicity & classification \\
  Hateful Memes & Image & toxicity & classification \\
  CHIME & Text & explanation & open-ended \\
  MemeBridge & Image & explanation & open-ended \\
  MemeBench & Image & diagnosis & open-ended \\
  \textbf{MemeX} & \textbf{Image} & \textbf{explanation} & \textbf{open-ended} \\
  \bottomrule
  \end{tabular}}
  \vspace{-0.4\baselineskip}
\end{wraptable}

\begin{wrapfigure}{r}{0.47\textwidth}
  \centering
  \vspace{-0.6\baselineskip}
  \includegraphics[width=\linewidth]{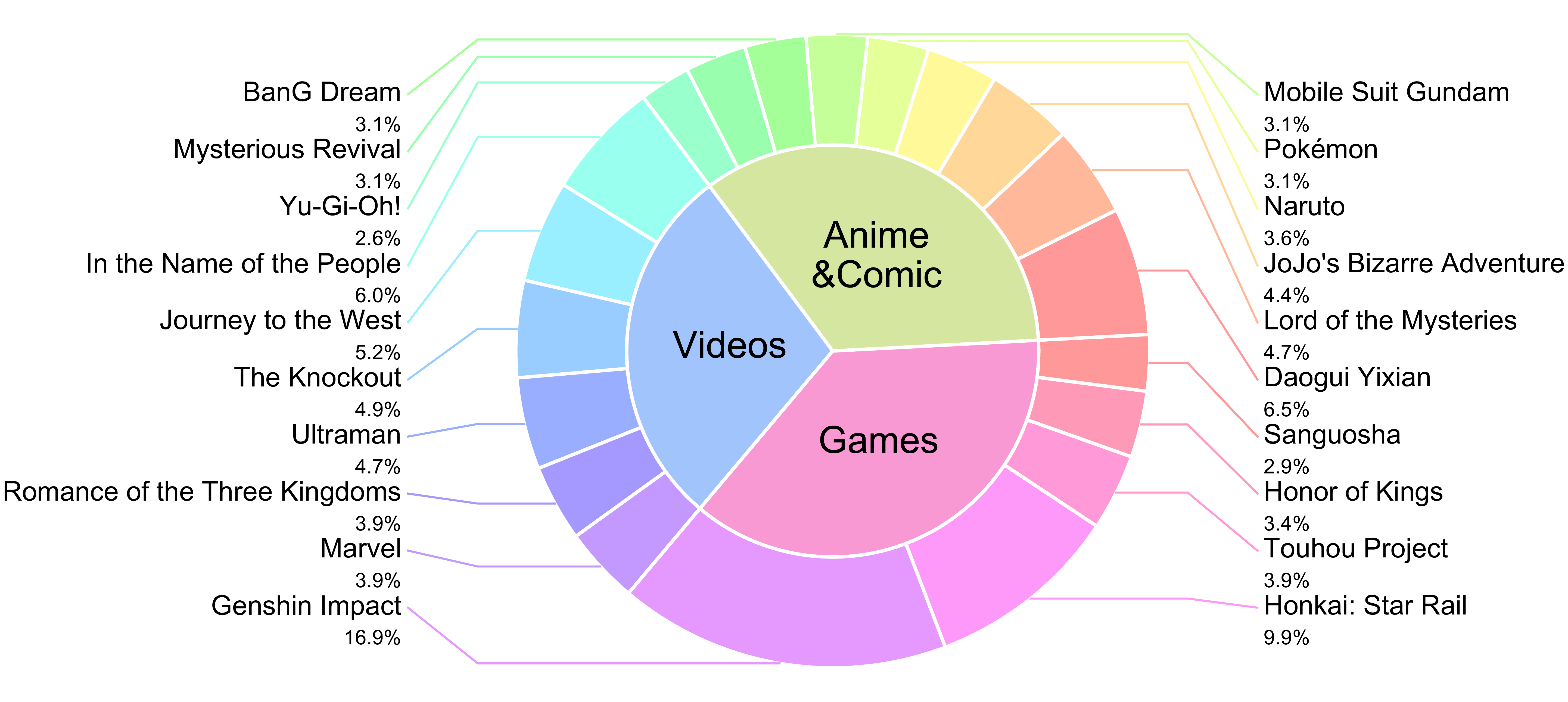}
  \caption{Category and intellectual-property distribution of MemeX. The benchmark spans nine content categories and combines globally familiar franchises with long-tail ACG properties.}
  \label{fig:data-chart}
  \vspace{-0.5\baselineskip}
\end{wrapfigure}

\paragraph{Benchmark.}
MemeX provides a controlled test bed for retrieval-intensive ACG meme explanation. It contains 1{,}000 image-based memes across nine categories and covers both globally recognized and long-tail properties. Knowledge-intensive categories form the majority, with 455 Anime \& Comic samples, 234 Games samples, and 85 Videos samples. Figure~\ref{fig:data-chart} reports the category and intellectual-property distribution, and Table~\ref{tab:dataset-comparison} positions MemeX against representative meme benchmarks.

The benchmark targets three recurring demands. Edited screenshots, image macros, comics, and mixed media introduce substantial visual and linguistic variation. Correct explanations often require characters, titles, events, and fandom references that are not recoverable from pixels alone. Overlaid text and culture-specific humor further require models to connect retrieved evidence rather than stop at entity lookup. These properties yield retrieval-dependent answers and heterogeneous tool procedures, including cases in which a native rollout group contains no accepted process.

\WFclear

\paragraph{Reference construction.}
Reference construction separates candidate drafting from expert verification. GPT-5 first drafts the visible entities, relevant source material, cultural background, and relation that produces the intended humor. Expert annotators then correct factual errors, add missing context and provenance cues, and finalize each reference through dual review and adjudication. The resulting reference supports evaluation of entity recognition, story understanding, and meme comprehension. It specifies the desired outcome but does not prescribe the tool trajectory that TraceBuilder must construct.

\paragraph{Models and adaptation.}
We evaluate Qwen3-VL-235B-A22B and Qwen3-VL-30B-A3B under a shared offline adaptation protocol. MemeX-100 is the standard stratified 100-sample adaptation split over English and non-English memes, with the latter primarily in Chinese, while evaluation uses the disjoint held-out set. Every method runs for three offline epochs with temperature $0.7$ and rollout size $G=8$. GPT-5 is the default curator. References are available to the judge and TraceBuilder during adaptation but are withheld at test time.

\input{sections/table_main_results}

The two partitions test whether adaptation relies on language-specific surface forms. The non-English partition contains localized phrases and community-specific references, whereas the English partition expresses globally shared franchises through different cultural conventions. We retain both scores in every main-table cell so that cross-language differences remain visible.

\paragraph{Evaluation and tools.}
Evaluation measures whether a generated explanation recovers the content represented in its expert reference. GPT-5 and Gemini3-Flash judge each response independently on character recognition, story understanding, and meme comprehension, assigning 0--5 points per dimension for a maximum score of 15. Character recognition covers entities and salient visual elements. Story understanding covers relations and contextual dependencies. Meme comprehension covers humor, irony, and cultural implications. A sensitivity study further varies curator and judge models. Tool access consists of Tavily/SERP text and image search with the top five results, script-validated region crops predicted by an LVLM, and Jina-embedding image-to-image retrieval.

\paragraph{Baselines.}
The comparison includes ReAct, ReAct with in-context examples, ReAct+ACE~\cite{zhang2025ace}, and ReAct+TF-GRPO~\cite{cai2025tfgrpo}. All methods use the same backbones, tools, adaptation data, and evaluation procedure. TF-GRPO can exploit relative differences within a failed group. TraceBuilder addresses the remaining case by supplying an accepted, outcome-correct process trace when every rollout is wrong.

\paragraph{Reproducibility.}
To keep the comparison controlled, all reported methods use the same decoding temperature, native rollout budget, tool interfaces, and held-out evaluation set. Main results retain both language partitions under both judges. Reported scores are averages over the fixed held-out examples for each condition. The appendix provides the judge prompt, learned guides, complete prompt templates, and full qualitative traces.

\paragraph{Supervision lifecycle.}
The supervision protocol separates outcome information from process evidence. Offline, the expert reference supplies the target for judging native trajectories and specifies the desired outcome when an all-failure group activates TraceBuilder. It never supplies tool observations. TraceBuilder must issue executable queries and ground the reconstructed process in the returned image or text evidence. A constructed trace enters the experience buffer only after it passes both the ordinary task-level judgment and the claim-level evidence check. Candidate guides are then screened for entity or answer leakage and compared with the preceding guides on held-out validation examples. At test time, the reference, experience buffer, curator state, and validation feedback are all unavailable.

\paragraph{Controlled comparison.}
The native rollout budget $G$ is identical across adaptive methods. TraceBuilder adds conditional construction only after all $G$ native attempts fail. The comparison therefore matches native exploration rather than total adaptation-time computation, since the conditional construction is the intervention being measured. ReAct retains the manual prompt, ReAct+ICL adds fixed demonstrations, ACE updates a shared context from execution feedback, and TF-GRPO uses within-group preferences. MemeMind keeps the same backbone and external interfaces, but constructs the missing accepted process and stores reusable acquisition rules in shared and tool-wise guides. We report every language--judge cell without selecting a favorable evaluator or averaging away cross-language variation.

%% file: sections/table_main_results.tex
\begin{table*}[t]
\caption{MemeX results. Scores are non-English / {\color{en_blue}English}; improvement is relative to the strongest competing method.}
\label{tab:main_results}
\centering
\footnotesize
\renewcommand{\arraystretch}{1.2}
\setlength{\tabcolsep}{3pt}
\resizebox{\textwidth}{!}{%
\begin{tabular}{@{} l cccc cccc @{}}
\toprule
\multicolumn{9}{c}{\cellcolor[gray]{0.96} \textbf{MemeX Performance: Non-English {\color{sep_gray}/} \color{en_blue}English-Language}} \\
\midrule
    & \multicolumn{4}{c}{\textbf{GPT-5 as Judge Model}} & \multicolumn{4}{c}{\textbf{Gemini3-Flash as Judge Model}} \\
\cmidrule(lr){2-5} \cmidrule(lr){6-9}
\textbf{Method} &
\textbf{Character Recog.} &
\textbf{Story Understanding} &
\textbf{Meme Comprehension} &
\textbf{Total} &
\textbf{Character Recog.} &
\textbf{Story Understanding} &
\textbf{Meme Comprehension} &
\textbf{Total} \\
\midrule
\multicolumn{9}{l}{\textit{Model: Qwen3-VL-235B-A22B}} \\
ReAct            & \mres{2.00}{2.10} & \mres{1.80}{1.85} & \mres{1.88}{1.90} & \mres{5.68}{5.85} & \mres{1.95}{2.05} & \mres{1.78}{1.80} & \mres{1.92}{1.95} & \mres{5.65}{5.80} \\
ReAct + ICL      & \mres{2.74}{2.80} & \mres{2.35}{2.40} & \mres{2.37}{2.45} & \mres{7.46}{7.65} & \mres{2.68}{2.75} & \mres{2.30}{2.35} & \mres{2.42}{2.50} & \mres{7.40}{7.60} \\
ReAct + ACE      & \mres{3.02}{3.10} & \mres{2.71}{2.78} & \mres{2.70}{2.75} & \mres{8.43}{8.63} & \mres{2.95}{3.02} & \mres{2.75}{2.80} & \mres{2.68}{2.72} & \mres{8.38}{8.54} \\
ReAct + TF-GRPO & \mres{3.07}{3.15} & \mres{2.89}{2.95} & \mres{2.66}{2.72} & \mres{8.62}{8.82} & \mres{3.02}{3.10} & \mres{2.92}{3.00} & \mres{2.60}{2.68} & \mres{8.54}{8.78} \\

\textbf{MemeMind} & \mresb{3.16}{3.25} & \mresb{3.13}{3.18} & \mresb{3.02}{3.10} & \mresb{9.32}{9.53} & \mresb{3.20}{3.28} & \mresb{3.15}{3.20} & \mresb{3.05}{3.12} & \mresb{9.40}{9.60} \\
\rowcolor[HTML]{F2F2F2}
\textit{Improv. \%} & \mres{+2.9\%}{+3.2\%} & \mres{+8.3\%}{+7.8\%} & \mres{+13.5\%}{+14.0\%} & \textbf{\mres{+8.1\%}{+8.0\%}} & \mres{+6.0\%}{+5.8\%} & \mres{+7.9\%}{+6.7\%} & \mres{+13.8\%}{+14.7\%} & \textbf{\mres{+10.1\%}{+9.3\%}} \\
\addlinespace[0.6em]
\multicolumn{9}{l}{\textit{Model: Qwen3-VL-30B-A3B}} \\
ReAct            & \mres{1.24}{1.30} & \mres{1.36}{1.40} & \mres{1.32}{1.35} & \mres{3.92}{4.05} & \mres{1.20}{1.25} & \mres{1.40}{1.45} & \mres{1.28}{1.30} & \mres{3.88}{4.00} \\
ReAct + ICL      & \mres{1.71}{1.78} & \mres{1.55}{1.62} & \mres{1.53}{1.60} & \mres{4.79}{5.00} & \mres{1.68}{1.75} & \mres{1.58}{1.65} & \mres{1.50}{1.55} & \mres{4.76}{4.95} \\
ReAct + ACE      & \mres{2.34}{2.42} & \mres{2.27}{2.35} & \mres{2.20}{2.28} & \mres{6.81}{7.05} & \mres{2.30}{2.38} & \mres{2.25}{2.32} & \mres{2.18}{2.25} & \mres{6.73}{6.95} \\
ReAct + TF-GRPO & \mres{2.33}{2.40} & \mres{2.47}{2.55} & \mres{2.21}{2.30} & \mres{7.01}{7.25} & \mres{2.28}{2.35} & \mres{2.50}{2.58} & \mres{2.18}{2.25} & \mres{6.96}{7.18} \\

\textbf{MemeMind} & \mresb{2.93}{3.02} & \mresb{2.69}{2.75} & \mresb{2.93}{3.01} & \mresb{8.55}{8.78} & \mresb{3.01}{3.08} & \mresb{2.75}{2.82} & \mresb{2.90}{2.98} & \mresb{8.66}{8.88} \\
\rowcolor[HTML]{F2F2F2}
\textit{Improv. \%} & \mres{+25.8\%}{+25.8\%} & \mres{+8.9\%}{+7.8\%} & \mres{+32.6\%}{+30.9\%} & \textbf{\mres{+22.0\%}{+21.1\%}} & \mres{+31.4\%}{+30.5\%} & \mres{+10.0\%}{+9.3\%} & \mres{+33.0\%}{+32.4\%} & \textbf{\mres{+24.4\%}{+23.7\%}} \\
\bottomrule
\end{tabular}%
}
\end{table*}

%% file: sections/5_experiments.tex
\FloatBarrier

\section{Experiments}
\label{experiments}

We evaluate MemeMind through four questions. \textbf{(Q1)} Does it outperform matched context-optimization baselines across backbones, languages, and judges? \textbf{(Q2)} What does each component contribute? \textbf{(Q3)} Do the gains persist across rollout budgets and choices of curator and judge? \textbf{(Q4)} Does adaptation produce the tool specialization that ToolGuide is intended to preserve?

\subsection{Main Results}
Table~\ref{tab:main_results} shows that MemeMind obtains the highest score in every evaluation cell. Under GPT-5 judging, it improves over TF-GRPO from 8.62/8.82 to 9.32/9.53 on Qwen3-VL-235B-A22B, corresponding to 8.1\%/8.0\%, and from 7.01/7.25 to 8.55/8.78 on Qwen3-VL-30B-A3B, corresponding to 22.0\%/21.1\%. Gemini3-Flash gives the same ordering. The gains cover all three dimensions and are larger on the smaller backbone, a pattern examined further in the ablation study.

The dimension-level results locate the main improvements. Under GPT-5, the 235B model exceeds TF-GRPO by 2.9\% in character recognition, 8.3\% in story understanding, and 13.5\% in meme comprehension. The corresponding gains for the 30B model are 25.8\%, 8.9\%, and 32.6\%. Thus, better factual recognition is accompanied by larger gains in the final interpretation. English examples exhibit the same pattern across localized and globally shared references.

\subsection{Ablation Study}
We isolate the roles of TraceBuilder and ToolGuide using cumulative ablations that progressively remove them from the full system.

\begin{table*}[t]
\centering
\begin{minipage}[t]{0.56\textwidth}
  \centering
  \captionof{table}{Cumulative ablation on the non-English partition with GPT-5 judging. ToolGuide is removed after TraceBuilder; $\Delta$ is relative to Full.}
  \label{tab:ablation_study}
  \vspace{0.3em}
  \scriptsize
  \renewcommand{\arraystretch}{1.16}
  \setlength{\tabcolsep}{3pt}
  \resizebox{\linewidth}{!}{%
  \begin{tabular}{lccccr}
  \toprule
  \textbf{Method} & \textbf{Character} & \textbf{Story} & \textbf{Meme} & \textbf{Total} & \textbf{$\Delta\%$} \\
  \midrule
  \multicolumn{6}{l}{\textit{Qwen3-VL-235B-A22B}} \\
  Full & 3.16 & 3.13 & 3.02 & 9.32 & - \\
  w/o TraceBuilder & 3.09 & 2.90 & 2.81 & 8.80 & $\downarrow$ 5.6\% \\
  w/o TraceBuilder \& ToolGuide & 2.91 & 2.75 & 2.80 & 8.48 & $\downarrow$ 9.0\% \\
  \midrule
  \multicolumn{6}{l}{\textit{Qwen3-VL-30B-A3B}} \\
  Full & 2.93 & 2.69 & 2.93 & 8.55 & - \\
  w/o TraceBuilder & 2.24 & 2.31 & 2.17 & 6.70 & $\downarrow$ 21.6\% \\
  w/o TraceBuilder \& ToolGuide & 2.13 & 2.17 & 2.16 & 6.46 & $\downarrow$ 24.4\% \\
  \bottomrule
  \end{tabular}}
\end{minipage}
\hfill
\begin{minipage}[t]{0.41\textwidth}
  \centering
  \captionof{table}{Cross-model curator--judge sensitivity on Qwen3-VL-30B-A3B. Bold marks the best curator for each judge.}
  \label{tab:cross-judge}
  \vspace{0.3em}
  \scriptsize
  \setlength{\tabcolsep}{3pt}
  \resizebox{\linewidth}{!}{%
  \begin{tabular}{lccc}
  \toprule
  \textbf{Curator} & \multicolumn{3}{c}{\textbf{Judge}} \\
  \cmidrule(lr){2-4}
   & GPT-5 & Gemini3-Flash & Qwen3-30B \\
  \midrule
  GPT-5 & 8.55 & \textbf{8.02} & \textbf{8.63} \\
  Gemini3-Flash & \textbf{8.61} & 7.53 & 7.71 \\
  Qwen3-30B & 6.84 & 6.91 & 7.27 \\
  \bottomrule
  \end{tabular}}
\end{minipage}
\end{table*}

The largest reduction follows the removal of TraceBuilder. Total score falls from 9.32 to 8.80 on the 235B model and from 8.55 to 6.70 on the 30B model, reductions of 5.6\% and 21.6\%. Removing ToolGuide as well lowers the scores to 8.48 and 6.46. This additional difference is concentrated in character and story scores, consistent with ToolGuide organizing evidence grounding and acquisition across tools.

\begin{figure*}[t]
  \centering
  \begin{minipage}[t]{0.48\textwidth}
    \centering
    \includegraphics[width=0.94\linewidth]{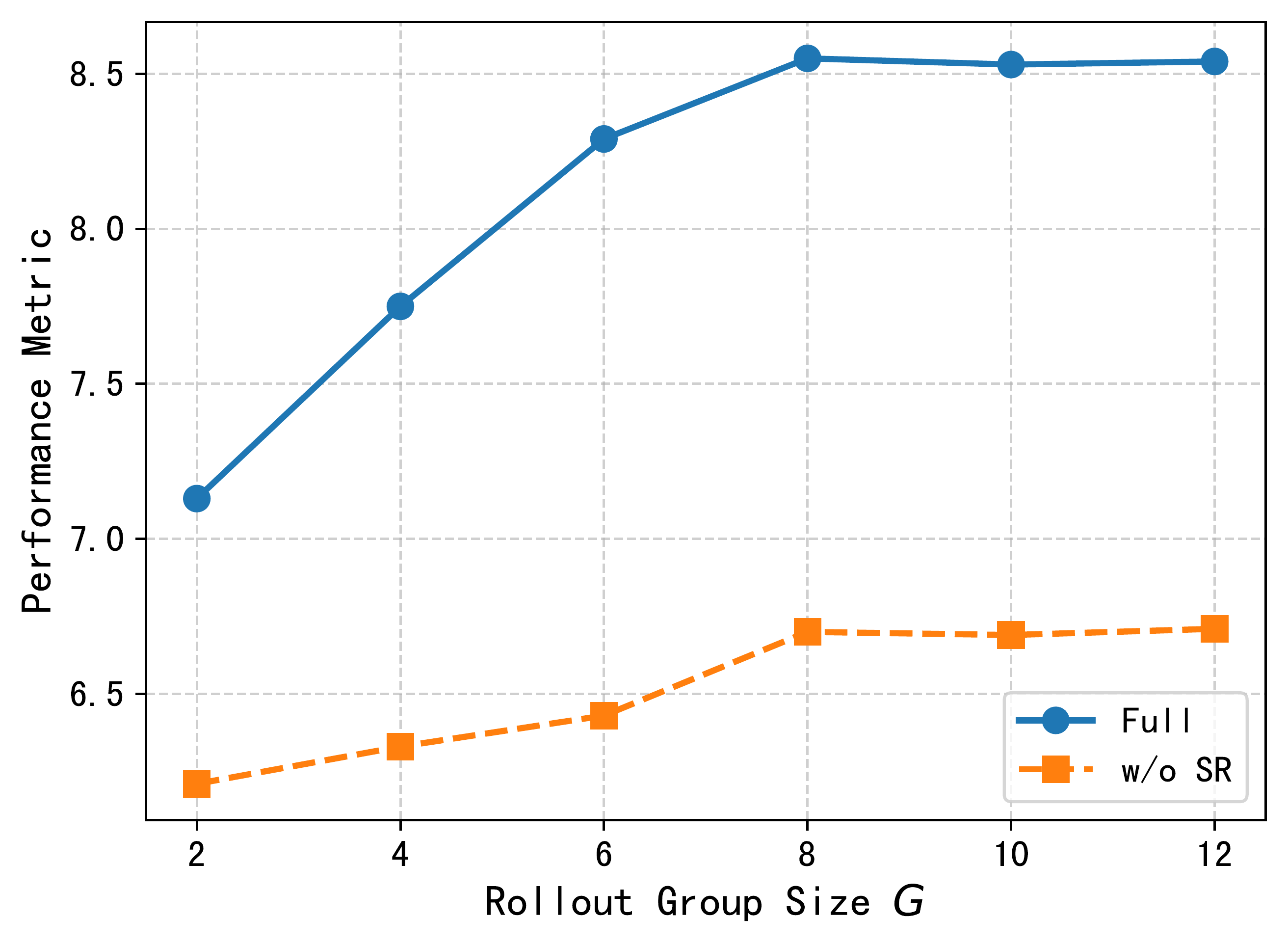}
    \captionof{figure}{Sensitivity to rollout group size $G$ on Qwen3-VL-30B-A3B. The full method remains above the variant without TraceBuilder from $G=2$ through $12$.}
    \label{fig:rollout_g}
  \end{minipage}
  \hfill
  \begin{minipage}[t]{0.48\textwidth}
    \centering
    \includegraphics[width=\linewidth]{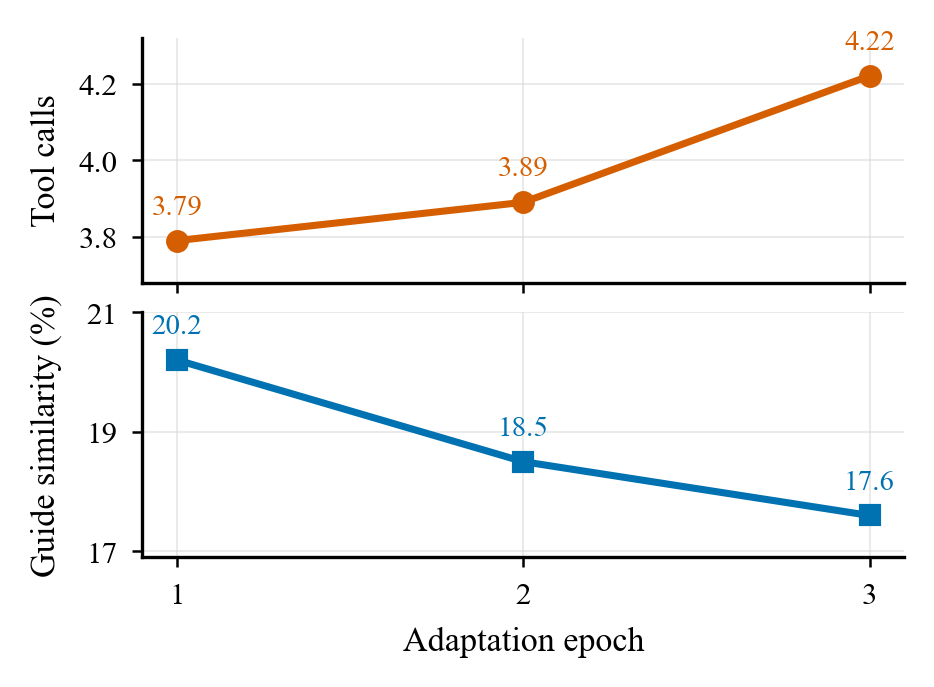}
    \captionof{figure}{Adaptation behavior on Qwen3-VL-30B-A3B. Tool use increases while cross-tool guide similarity decreases, consistent with increasingly specialized acquisition procedures.}
    \label{fig:adaptation-behavior}
  \end{minipage}
\end{figure*}

\textbf{Rollout group size.} Figure~\ref{fig:rollout_g} shows that the full method remains above the variant without TraceBuilder even at $G=2$. Performance increases as exploration grows, approaches saturation near $G=8$, and remains stable through $G=12$.

\subsection{Robustness to Evaluator Choice}

Table~\ref{tab:cross-judge} examines whether the result depends on a particular pairing of curator and judge. The strongest curator varies by judge. Gemini3-Flash reaches 8.61 under GPT-5 judging, whereas GPT-5 reaches 8.63 under the Qwen3-30B judge. Every curator produces a substantial score under each of the three judges, and the ordering is not tied to self-evaluation. Together with Figure~\ref{fig:rollout_g}, these results indicate that the observed gain persists when either the exploration budget or the auxiliary models used to update and evaluate guides are changed.

\subsection{What Changes During Adaptation?}

Figure~\ref{fig:adaptation-behavior} reports two behavioral measurements over three offline epochs. Average tool calls increase from 3.79 to 4.22, while ROUGE-L similarity among tool-wise guides decreases from 20.2\% to 17.6\%. These opposing trends are consistent with the intended division of labor. Adaptation encourages more evidence acquisition while making the instructions for image retrieval, text search, and visual grounding less interchangeable. The trends do not by themselves imply that every additional call is useful. Their relevance comes from appearing together with higher task scores and lower cross-tool similarity, which is inconsistent with merely repeating the same undirected search behavior. Taken together, the quantitative results show a consistent pattern. Table~\ref{tab:main_results} gives the overall improvement, Table~\ref{tab:ablation_study} identifies TraceBuilder as the main contributor, Figure~\ref{fig:rollout_g} shows that the advantage is not limited to one rollout budget, and Table~\ref{tab:cross-judge} shows that it is not explained by a single curator and judge pairing.

\subsection{Data Efficiency of Context Adaptation}

Figure~\ref{fig:data-efficiency} varies the adaptation set from 50 to 200 examples on Qwen3-VL-30B-A3B. For this sensitivity analysis, we form nested stratified subsets while keeping the held-out evaluation set fixed.

\noindent
\begin{minipage}[t]{0.49\textwidth}
  \vspace{0pt}
  The 100-example point is the standard MemeX-100 setting used elsewhere. Total score increases from 8.43 with 50 examples to 8.55, with most of the improvement obtained by approximately 100 examples. A compact adaptation set therefore appears sufficient to expose reusable decisions about region selection, query construction, tool ordering, and evidence integration. Beyond 100 examples, performance remains stable as the guide approaches its strongest observed score. This saturation suggests that recurring acquisition patterns matter more than repeated exposure to similar cases. It also makes the approach practical when expert references are costly to construct.
\end{minipage}
\hfill
\begin{minipage}[t]{0.47\textwidth}
  \vspace{0pt}
  \centering
  \includegraphics[width=0.96\linewidth]{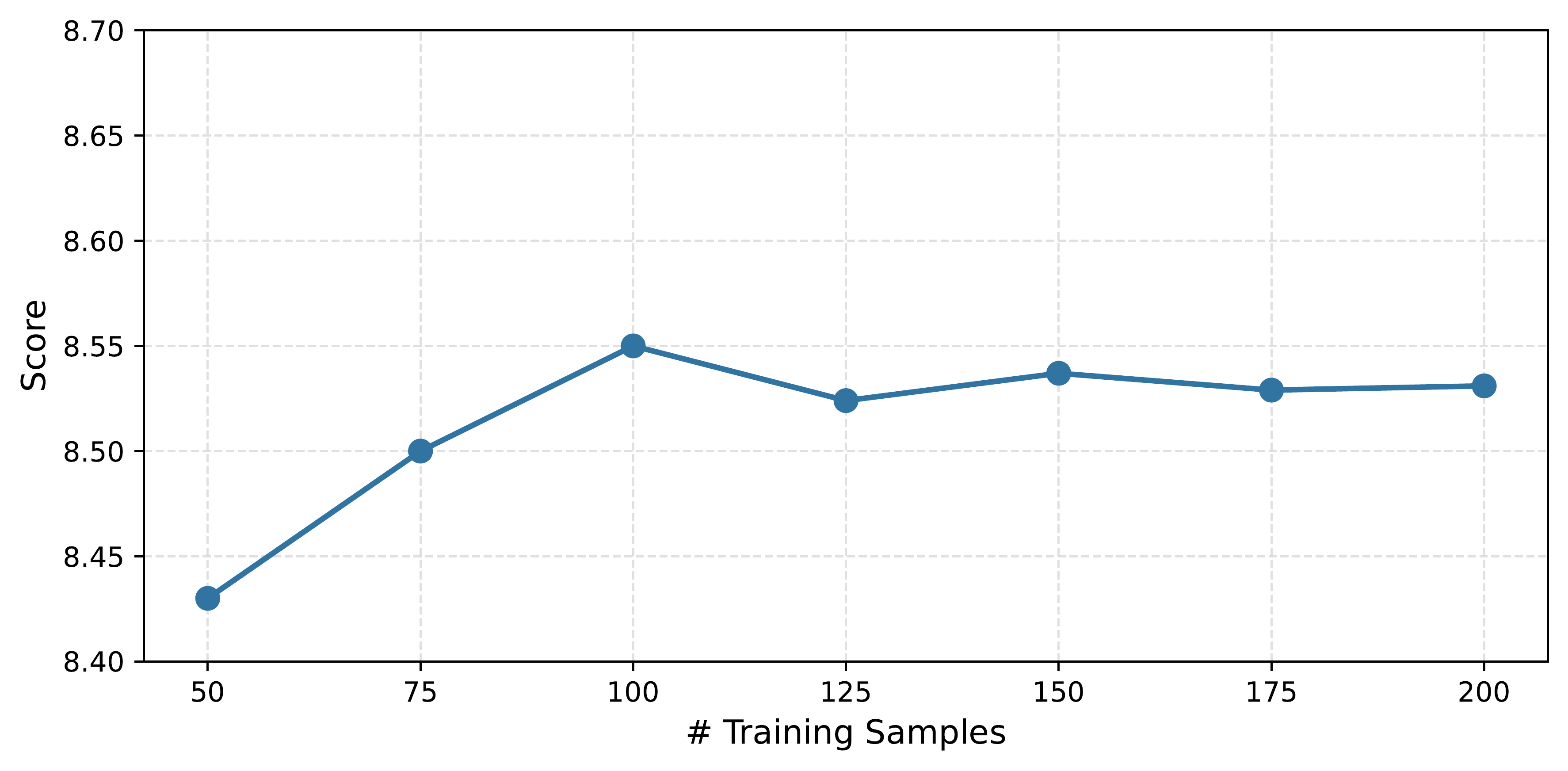}
  \captionsetup{hypcap=false}
  \captionof{figure}{Data-efficiency scaling on Qwen3-VL-30B-A3B. Performance peaks near 100 adaptation examples and remains stable through 200.}
  \label{fig:data-efficiency}
\end{minipage}

\subsection{Process-Level Evidence}

Aggregate scores show the effect of adaptation but not how the learned guides alter evidence acquisition. Table~\ref{tab:process-evidence} summarizes a held-out execution whose answer depends on both a Western television character and a Chinese film reference. Both runs use the same input and tools, with only the deployed guides changed.

\begin{table}[ht]
\caption{A held-out trace contrast. Search-result pages are omitted for space; both runs use the same input and tool interfaces.}
\label{tab:process-evidence}
\centering
\scriptsize
\setlength{\tabcolsep}{2.5pt}
\renewcommand{\arraystretch}{1.08}
\begin{tabularx}{\columnwidth}{@{}p{0.17\columnwidth}XX@{}}
\toprule
\textbf{Decision} & \textbf{Original prompt} & \textbf{MemeMind guides} \\
\midrule
Initial belief
& Commits early to Tom Welling and Andy Lau from facial resemblance.
& Splits the three panels and keeps names tentative until retrieval supports them. \\
Acquisition
& After conflicting image results, issues generic ``superpower decay meme'' queries.
& Confirms Homelander visually, then hands the unresolved phrase ``I don't eat beef'' to text search. \\
Final account
& Leaves two identities unresolved and gives a generic decline-from-heroism reading.
& Recovers Cao Shaolin from \emph{Call of Heroes} and explains idealism $\rightarrow$ authoritarian corruption $\rightarrow$ nihilistic violence. \\
\bottomrule
\end{tabularx}
\end{table}

The original prompt reacts to contradictory image results but does not retain the contradiction as a search target. Its subsequent generic queries recover the template while missing the culturally decisive identities. MemeMind instead searches panel-level regions, treats visual matches as hypotheses, and transfers an exact unresolved phrase from image evidence to text search. This recovers the identity and source context needed for the final interpretation. The case illustrates the two trends in Figure~\ref{fig:adaptation-behavior}. Increased tool use is paired with differentiated responsibilities rather than repeated calls to interchangeable tools.

\paragraph{What the guides retain.}
The final artifacts retain procedures rather than case answers. Table~\ref{tab:guide-content} shows that the two retrieval tools use different invocation conditions, query construction rules, evidence tests, and handoff objects. The shared guide then connects their outputs to the final explanation.

\begin{table}[t]
\caption{Condensed content of the learned tool-wise guides. Full instructions and the complete case trace are in the appendix.}
\label{tab:guide-content}
\centering
\footnotesize
\setlength{\tabcolsep}{3pt}
\renewcommand{\arraystretch}{1.12}
\begin{tabularx}{\columnwidth}{@{}p{0.18\columnwidth}>{\raggedright\arraybackslash}X>{\raggedright\arraybackslash}X@{}}
\toprule
\textbf{Decision} & \textbf{Image retrieval} & \textbf{Text search} \\
\midrule
Invocation & Unresolved entity, scene, or salient region & Grounded hypothesis or exact phrase \\
Query & Full image, then panel/ROI and discriminative attributes & Source, character attributes, and quoted text \\
Acceptance & Multiple observable features and independent matches & OCR-derived names remain tentative until supported \\
Handoff & Neutral descriptors and regions & Source-backed identity and context \\
\bottomrule
\end{tabularx}
\end{table}

TraceBuilder receives the expert reference only for offline adaptation pairs whose rollouts all fail, and it uses the same search interfaces as the native rollouts. References, traces, curator state, and judge feedback are discarded after adaptation. In the held-out case above, the frozen backbone receives only the meme, final guides, and ordinary tools.

\paragraph{Trace-level comparison.}
With the original prompt, the first round identifies the upper panel as Henry Cavill's Superman but guesses Tom Welling and Andy Lau for the remaining panels based on facial familiarity. Separate image searches contradict these guesses. The second panel points toward \emph{The Boys}, while the third repeatedly co-occurs with the Chinese phrase ``I don't eat beef.'' The model notices the mismatch, but then issues two broad template queries, ``if you had superpowers meme'' and ``superpower decay meme Chinese.'' They recover the generic three-stage template but not the missing sources. The final explanation therefore describes a decline from heroic responsibility to convenience or play while leaving the two culturally decisive identities unresolved.

The MemeMind run changes three decisions instead of simply adding search calls. First, it crops the panels into separate regions and excludes the overlaid day labels, reducing interference across visual units. Second, it keeps retrieved identities provisional. Image retrieval supports Superman and Homelander but does not force a name for the last panel. Third, it retains the unresolved phrase as an evidence handoff and submits it verbatim to text search. The returned source links the phrase to Cao Shaolin in the 2016 film \emph{Call of Heroes}. The final explanation can then ground the intended escalation. Superman represents idealized heroic power, Homelander its narcissistic and authoritarian corruption, and Cao Shaolin arbitrary violence without moral constraint. The phrase functions as both a retrieval key and cultural shorthand for capricious brutality.

\paragraph{Process interpretation.}
The deployed guides change three directly observable decisions. These are the unit of visual search, the timing of identity commitment, and the object passed from image retrieval to text search. This behavior connects the held-out trace to the aggregate adaptation dynamics. More tool calls reflect additional targeted evidence acquisition, while lower cross-tool guide similarity reflects a clearer division of labor among the interfaces. The trace thus provides a process-level account of the improvements observed in the held-out results and ablations.

\paragraph{Connecting TraceBuilder and ToolGuide.}
This held-out run contains no expert reference and does not invoke TraceBuilder. It instead shows the type of procedure retained by ToolGuide after offline adaptation. TraceBuilder operates upstream. For an all-failure adaptation query, it attempts to construct an accepted execution containing region selection, targeted queries, returned observations, and links between evidence and claims. ToolGuide abstracts these decisions across the buffer, while its leakage audit removes case-specific answers. The two components therefore serve distinct roles. Constructed traces supply the missing positive process during adaptation, and the final guides are the only learned artifact used at inference.

\paragraph{Component-level evidence.}
Table~\ref{tab:ablation_study} identifies TraceBuilder as the main source of the component-level gain. The further cumulative reduction after removing ToolGuide shows that structured guide learning contributes beyond repaired traces. The guide content, adaptation dynamics, and held-out execution clarify this additional contribution through differentiated queries, acceptance tests, and evidence handoffs across tools.

%% file: sections/6_conclusion.tex
\section{Conclusion}
\label{conclusion}

We study all-failure rollout groups as a missing-process problem in offline context optimization. An outcome reference may be available even when ordinary exploration produces no accepted account of how heterogeneous tools can reach it. \method{} addresses this gap by separating process construction from process consolidation. TraceBuilder converts the offline reference into a plan, executes the available tools, verifies the evidence, and admits the trace only when it passes the task and evidence checks. ToolGuide consolidates the collected traces into a shared guide and interface-specific procedures for a frozen test-time agent. Experiments yield three main findings. Addressing all-failure experience provides the largest component-level gain, particularly for the smaller backbone. The improvement persists across language partitions, judge models, curator and judge pairings, and rollout budgets. Adaptation also produces reusable procedural structure. Tool use becomes more active, tool-wise guides become more differentiated, a compact adaptation set reaches strong performance, and held-out traces show targeted evidence handoffs across visual grounding, image retrieval, and text search. These results indicate that reference-guided trace construction can turn outcome supervision into reusable process supervision for tool-augmented multimodal agents.

%% file: sections/7_impact.tex
\section{Societal Impact and Data Considerations}

\method{} examines retrieval-supported interpretation of visual internet culture. Improved evidence acquisition may help systems explain cross-lingual references and reduce unsupported guesses about entities. However, meme data can include copyrighted characters, user-created images, offensive language, and culturally sensitive material. MemeX is used for research evaluation. The benchmark records provenance cues during expert review, and the method neither infers private attributes nor identifies ordinary individuals. Deployment beyond this setting should preserve source attribution, respect content-removal requests and platform policies, and avoid presenting uncertain cultural interpretations as facts. The tool-wise guides retain uncertainty when external evidence is insufficient, which helps limit confident misidentification without eliminating the underlying risk.

%% file: appendix/appendix.tex
\section{Prompt Architecture and Complete Artifacts}
\label{prompts-utilized-in-mememind}

This section records the prompts used at each stage of \textsc{MemeMind}. Figure~\ref{fig:default-guidance} gives the \textbf{Default Guidance} and few-shot reasoning examples used to initialize the agent. The seed protocol specifies how web search, image grounding, and image-to-image retrieval are coordinated over visual, textual, and cultural evidence.

During offline adaptation, parallel rollouts and judge feedback provide native execution experience. When cultural ambiguity or visual complexity causes an all-failure group, \textbf{TraceBuilder} uses the outcome reference to construct a positive process trace. \textbf{ToolGuide} then updates a shared guide together with tool-specific guides, retaining separate acquisition procedures within one interpretation workflow.

The remaining artifacts follow this execution order:
\begin{itemize}
    \item The \textbf{experience extraction system prompt} maintains an ordered library of ideal think--act trajectories. From the judge-derived quality signal, it selects one operation such as Add, Modify, Delete, or No Change (Figure~\ref{fig:exp-extract-prompt});

    \item The \textbf{experience extraction template} supplies the previous examples, trajectory summaries, and reference answers, and requires a constrained single-step update that returns the complete revised set (Figure~\ref{fig:exp-extract-template});

    \item The \textbf{judgment system prompt} scores entity recognition, story and context understanding, and relational or meme comprehension, and returns the reasons for each score (Figure~\ref{fig:judge-prompt});

    \item The \textbf{task prompt} defines the interaction format, round limit, tool-call constraints, and the reasoning required before search or final interpretation (Figure~\ref{fig:full-agent-prompt}).
\end{itemize}

\begin{figure*}[htbp]
    \centering
    \begin{tcolorbox}[
        colback=white,
        colframe=black,
        title=\textbf{EXP\_EXTRACT\_SYSTEM\_PROMPT},
    ]
    \noindent \textbf{[Role Definition]} \\
    You are an \textbf{AI Expert} specializing in summarization, generalization, and iterative optimization. Your core objective is to maintain and refine an ordered library of \textit{ideal think-act trajectories} (few-shot examples) to guide a meme-understanding agent.

    \noindent \textbf{[Input Components]} \\
    $\bullet$ \texttt{prev\_fewshots}: The current list of few-shot examples. \\
    $\bullet$ \texttt{trajectories\_summary}: Execution summary of a new meme interpretation, including a \textbf{judge-derived trajectory quality signal} for each trajectory. \\
    $\bullet$ \texttt{reference\_answer}: The ground-truth interpretation of the new meme.

    \noindent \textbf{[Optimization Logic]} \\
    Read the \textbf{trajectory quality signal} to determine the necessary refinement action:
    \begin{itemize}[leftmargin=1.5em, itemsep=2pt, label=--]
        \item \textbf{Accepted trajectory (clears the acceptance bar):} Effective trajectory. Action: \texttt{[Add]} to library or \texttt{[Modify]} a suboptimal example.
        \item \textbf{Rejected trajectory (below the acceptance bar):} Reflect on the \texttt{reference\_answer}, reconstruct a reference-guided ideal trajectory, and \texttt{[Add]} or \texttt{[Modify]}.
    \end{itemize}

    \noindent \textbf{[Operational Constraints]} \\
    1. \textbf{Single-Step Edit}: Perform exactly \textbf{one} action per iteration: \texttt{[Add]}, \texttt{[Modify]}, \texttt{[Delete]}, or \texttt{[No Change]}. \\
    2. \textbf{Library Limit}: The list must contain a maximum of \textbf{5} examples. \\
    3. \textbf{Persistence}: Unchanged examples must remain intact.

    \noindent \textbf{[Output Specifications]} \\
    -- \textbf{Format}: Strictly follow: \textit{``Think: [Reasoning] $\rightarrow$ Act: [Tool] $\rightarrow \dots$''} \\
    -- \textbf{Encapsulation}: Enclose the updated list within an updated-fewshots block delimited by the designated tags.
    \end{tcolorbox}
    \caption{System Prompt for the Experience Extraction.}
    \label{fig:exp-extract-prompt}
\end{figure*}

\begin{figure*}[htbp]
    \centering
    \begin{tcolorbox}[
        colback=white, colframe=black,
        title=EXP\_EXTRACT\_TEMPLATE,
    ]

    \textbf{[Task]} \\
    Your task is to iteratively optimize the \texttt{[Few-shot Examples List]} based on the following inputs:
    \begin{itemize}
        \item \texttt{[Meme Analysis Agent's Trajectory and Thought Process Summary]}
        \item \texttt{[Meme]}
        \item \texttt{[Correct Answer for the Meme]}
    \end{itemize}
    Finally, output an \texttt{[Updated Few-shot Examples List]} (Maximum 10 examples).

    \textbf{[Context Inputs]}
    \begin{quote}
        \textbf{Previous Few-shots:} \{\{ few\_shots \}\} \\
        \textbf{Trajectory Summary:} \{\{ trajectories\_summary \}\} \\
        \textbf{Reference Answer:} \{\{ reference\_answer \}\} \\
        \textbf{Target Meme:} [Meme Content]
    \end{quote}

    \textbf{[Optimization Rules]}
    \begin{enumerate}
        \item For each optimization, choose exactly \textbf{one} action: \texttt{[Add]}, \texttt{[Modify]}, or \texttt{[Delete]}.
        \item Operation must be performed on a \textbf{single} instruction/example per iteration.
        \item All other instruction points must remain unchanged.
        \item Generate and output the \textbf{complete updated ordered list}.
    \end{enumerate}

    \textbf{[Output Format]} \\
    Please follow the thought process, then place the final updated complete ordered list of examples inside a dedicated updated-fewshots block delimited by the designated opening and closing tags.

    \end{tcolorbox}
    \caption{Template for the Experience Extraction.}
    \label{fig:exp-extract-template}
\end{figure*}

\begin{figure*}[htbp]
    \centering
    \begin{adjustbox}{max width=\textwidth,max totalheight=0.86\textheight,center}
    \begin{minipage}{\textwidth}
    \begin{tcolorbox}[
        colback=white,
        colframe=black,
        title=JUDGE\_PROMPT,
    ]

    You are a strict and impartial meme image analysis reviewer. Based on the ``reference answer,'' please score the ``model response'' on the following three dimensions on a scale of 0--5, and provide detailed reasoning.

    \vspace{10pt}
    \textbf{[Scoring Criteria]}

    \begin{enumerate}
        \item \textbf{Entity Recognition Score (entity\_score) -- 0 to 5 points}: Measures whether the model correctly and comprehensively identifies key entities in the meme (e.g., people, characters, objects, scenes, text, etc.).
        \begin{itemize}
            \item \textbf{5 points (Perfect Identification)}: Accurately identifies all key entities, fully consistent with the reference answer, with no omissions or errors.
            \item \textbf{4 points (Mostly Accurate)}: Identifies the vast majority of key entities, but misses a few minor ones or has slight errors.
            \item \textbf{3 points (Partially Accurate)}: Identifies core entities but omits multiple key entities or makes clear errors.
            \item \textbf{2 points (Minimally Accurate)}: Only identifies a few non-core entities, or misidentifies core entities.
            \item \textbf{1 point (Nearly Inaccurate)}: Fails to correctly identify almost any entities.
            \item \textbf{0 points (Completely Wrong)}: Identifies no entities at all.
        \end{itemize}

        \item \textbf{Story/Context Recognition Score (story\_score) -- 0 to 5 points}: Measures whether the model understands the background story, plot, original work, or cultural context.
        \begin{itemize}
            \item \textbf{5 points (Deep Understanding)}: Fully and accurately grasps the background story/cultural context, consistent with the reference answer.
            \item \textbf{4 points (Basic Understanding)}: Correctly understands the core background but has minor inaccuracies or omissions regarding secondary details.
            \item \textbf{3 points (Superficial Understanding)}: Understands the general framework but fails to capture key plot points or deeper background.
            \item \textbf{2 points (Fragmented Understanding)}: Mentions only scattered fragments without forming a coherent contextual understanding.
            \item \textbf{1 point (Misunderstood)}: Fundamentally misunderstands the background story.
            \item \textbf{0 points (No Understanding)}: Fails to recognize any relevant background story or context.
        \end{itemize}

        \item \textbf{Entity Relationship / Meme Comprehension Score (relation\_score) -- 0 to 5 points}: Measures whether the model accurately captures the interaction logic and understands the humor, irony, or metaphor.
        \begin{itemize}
            \item \textbf{5 points (Precise Interpretation)}: Fully and accurately captures relationships and deeply understands the humor, irony, or metaphor.
            \item \textbf{4 points (Generally Correct Interpretation)}: Correctly understands the main punchline, but slightly misses deeper implications.
            \item \textbf{3 points (Surface-Level Interpretation)}: Understands direct relationships but fails to grasp the core humor or satirical intent.
            \item \textbf{2 points (Vague Interpretation)}: Only vaguely senses some relationship but cannot clearly articulate the joke.
            \item \textbf{1 point (Incorrect Interpretation)}: Fundamentally misinterprets the relationship between entities.
            \item \textbf{0 points (No Comprehension)}: Completely fails to understand the meaning of the meme.
        \end{itemize}
    \end{enumerate}

    \textbf{[Output Format]} \\
    Return a step-by-step \texttt{reasons} field explaining the scoring, followed by integer-valued \texttt{entity\_score}, \texttt{story\_score}, and \texttt{relation\_score} fields, each delimited by the designated tags.

    \end{tcolorbox}
    \end{minipage}
    \end{adjustbox}
    \caption{System Prompt for the Judgement.}
    \label{fig:judge-prompt}
\end{figure*}

\begin{figure*}[htbp]
    \centering
    \begin{adjustbox}{max width=\textwidth,max totalheight=0.86\textheight,center}
    \begin{minipage}{\textwidth}
    \begin{tcolorbox}[
        colback=white,
        colframe=black,
        title=\texttt{TASK\_PROMPT},
    ]
    You are a top-tier Meme Interpretation Agent. Explain the input meme deeply and efficiently with minimal steps.

        \textbf{[Core Rules]}
        \begin{enumerate}
            \item \textbf{One tool per round}: Only one text-search or image-search call is allowed per turn.
            \item \textbf{Mandatory two-part output}: Every response MUST consist of exactly:
            \begin{itemize}
                \item First: a \texttt{reasoning} block.
                \item Second: \textbf{exactly one} of a text-search, image-search, or conclusion block.
            \end{itemize}
            \item \textbf{Stop when ready}: Once sufficient, output the \texttt{reasoning} block plus a \texttt{conclusion} block---no tools.
            \item \textbf{Maximum five rounds}: You must reach the \texttt{conclusion} block by the fifth interaction at the latest.
            \item \textbf{NEVER output plain text}: Any content outside the four designated blocks (reasoning, text-search, image-search, conclusion) is strictly forbidden.
        \end{enumerate}

        \textbf{[Tool Formats]}

        Text-search block payload:
        \begin{verbatim}
        {"queries": ["string1", "string2", ...]}
        \end{verbatim}

        Image-search block payload:
        \begin{verbatim}
        {
          "rois": [
            [x_min, y_min, x_max, y_max]
          ]
        }
        \end{verbatim}

        -- Coordinates: [0.00, 1.00], two decimal places.

        \textbf{[Output Instructions]}
        \begin{itemize}
            \item \textbf{Exploration round}: In the \texttt{reasoning} block, identify knowledge gaps and justify the next search; then emit a text-search or image-search block.
            \item \textbf{Final round}: In the \texttt{reasoning} block, explain how evidence supports your interpretation; then emit a \texttt{conclusion} block with the full explanation---meaning, origin, cultural context, key elements.
        \end{itemize}
    \end{tcolorbox}
    \end{minipage}
    \end{adjustbox}
    \caption{Task Prompt.}
    \label{fig:full-agent-prompt}
\end{figure*}

\begin{figure*}[htbp]
    \centering
    \begin{adjustbox}{max width=\textwidth,max totalheight=0.86\textheight,center}
    \begin{minipage}{\textwidth}
    \begin{tcolorbox}[
        colback=white,
        colframe=black,
        title=\texttt{DEFAULT\_GUIDANCE\_AND\_EXAMPLES},
    ]
    \noindent \textbf{[Default Guidance and Examples]} \\
    \rule{\linewidth}{0.2pt} \\
    \textbf{1. \texttt{DEFAULT\_GUIDANCE\_TEXT}} \\
    -- Write highly detailed and direct search \texttt{queries}. Queries should be more targeted and specific. \\
    \textbf{2. \texttt{DEFAULT\_GUIDANCE\_IMAGE}} \\
    -- Write specific character-related \texttt{queries} to search for images. Queries should be targeted and specific. Compare the retrieved images with the meme content for analysis. \\
    \textbf{3. \texttt{DEFAULT\_GUIDANCE\_FINAL\_ANALYSIS}} \\
    -- Search-Analyze-Summarize: Extract key perspectives and core conclusions.

    \vspace{1em}
    \noindent \textbf{[\texttt{DEFAULT\_FEW\_SHOT\_EXAMPLES}]} \\
    \rule{\linewidth}{0.2pt} \\
    \textbf{Example 1:} \\
    \textbf{Observe Image} $\rightarrow$ \textbf{Think}: The characters in the image seem to be making a certain comparison. $\rightarrow$ \textbf{Tool}: \texttt{search\_image}(``characters in the image'') $\rightarrow$ \textbf{Think}: Search results indicate the character is ``Character A'' from the anime \textit{JoJo's Bizarre Adventure}. $\rightarrow$ \textbf{Tool}: \texttt{search\_text}(``JoJo Character A story'') $\rightarrow$ \textbf{Think}: Results show Character A's backstory is relevant to the meme; attempting to interpret. $\rightarrow$ \textbf{Conclusion}: This meme expresses...
    \end{tcolorbox}
    \end{minipage}
    \end{adjustbox}
    \caption{Default Guidance and Few-Shot reasoning examples.}
    \label{fig:default-guidance}
\end{figure*}

\section{MemeMind Learned Protocols}
\label{app:protocols}

This section reports representative text-search and image-search protocols produced during offline optimization. The two protocols expose how \textsc{MemeMind} separates tool-specific acquisition rules while coordinating them for ACG meme understanding.

Figure~\ref{fig:guidance-text-search} gives the learned \textbf{text-search} protocol. Its purpose is to prevent an early OCR guess or speculative caption from fixing the interpretation. The protocol first enumerates observable visual attributes and validates candidate identities through image evidence. It then forms text queries, prioritizes authoritative sources such as official notes, developer forums, and curated wikis, and records uncertainty so that each query can be revised when later evidence conflicts.

Figure~\ref{fig:guidance-image-search} gives the learned \textbf{image-search} protocol. It decomposes reverse image retrieval into full-image and region-of-interest searches over general results and specialized template or still-image databases. Composite memes are separated into visual components, while small UI elements, symbols, and weapon shapes provide discriminative evidence. Character identities require both facial or sprite similarity and corroborating non-facial evidence from independent sources.

Together, the protocols define a tool-wise acquisition process. Text search resolves source context and cultural relations, image search grounds visual identity, and cross-channel validation prevents unsupported attribution. The complete protocols below are reproduced without modification.

\begin{figure*}[htbp]
    \centering
    \begin{adjustbox}{max width=\textwidth,max totalheight=0.86\textheight,center}
    \begin{minipage}{\textwidth}
    \begin{tcolorbox}[
        colback=white,
        colframe=black,
        title=\textbf{Protocol for search\_text},
    ]
    \noindent \textbf{1. Image-First Query Construction to Avoid Anchoring Bias:} \\
    Prior to constructing highly targeted text queries, perform template/scene confirmation based on visual elements.
    \begin{itemize}[leftmargin=1.5em, itemsep=2pt, label=--]
        \item Structurally list observable visual elements (gender, hair color, clothing details/emblems, weapon/prop shapes, facial expressions, UI/skill icons, on-image text/dialogue bubbles).
        \item Immediately execute a fast image search for the full image and key ROIs (panels/crops).
        \item Use OCR or visual guesses as secondary leads only if image search fails to identify a significant template.
        \item If image search yields high-confidence template/film candidates (similarity $\ge$ threshold), prioritize these as primary keywords for text queries.
        \item Cross-validate OCR results: If OCR suggests entities (names, titles), trigger image searches to verify against official data. Generate variants for homophones/near-synonyms and perform parallel searches. Mark OCR names as ``tentative/unverified'' unless supported by $\ge 2$ independent sources.
    \end{itemize}

    \vspace{0.6em}
    \noindent \textbf{2. Cross-Domain Origin Tracking:} \\
    Beyond social platforms, search authoritative sources: official patch notes, developer announcements, game wikis, and primary patch analysis (e.g., Bilibili/Reddit/Forums). Filter by version number/date and merge OCR text into queries to lock down context.

    \vspace{0.6em}
    \noindent \textbf{3. Template \& Scene Keyword Management:} \\
    Maintain a candidate list of common meme templates. Cross-reference template names with character/UI keywords (e.g., \textit{``Genshin Raiden Shogun distracted boyfriend''}) and record hit context/interpretative comments.

    \vspace{0.6em}
    \noindent \textbf{4. Precise OCR \& Filter Utilization:} \\
    Use exact OCR text for verbatim matching. Apply site filters (\texttt{site:pixiv}, \texttt{site:reddit}, etc.) and time ranges. Verify authorship or source labels (signatures, post descriptions, or explanatory comments).

    \vspace{0.6em}
    \noindent \textbf{5. Query Logging \& Uncertainty Handling:} \\
    Record high-confidence queries to reduce redundant calls. For multiple interpretations, list candidate sources and score them (High/Medium/Low confidence). Generate clarifying questions to resolve uncertainties (e.g., \textit{``Is `X' in the image an OCR error or a pun?''}).

    \vspace{0.6em}
    \noindent \textbf{6. Meme-Centric Query Construction:} \\
    Avoid overly generic queries that treat the entire image as a single unit (e.g., ``Character1 Character2 meme''). Instead, decompose the image into its constituent meme components and search for each recognizable sub-meme or trope individually (e.g., ``distracted boyfriend meme'', ``drake posting format'', ``this is fine dog reaction''). Prioritize well-established meme templates and their canonical keywords to improve retrieval precision.
    \end{tcolorbox}
    \end{minipage}
    \end{adjustbox}
    \caption{Learned protocol for text search.}
    \label{fig:guidance-text-search}
\end{figure*}

\begin{figure*}[htbp]
    \centering
    \begin{tcolorbox}[
        colback=white,
        colframe=black,
        title=\textbf{Protocol for search\_image},
        fonttitle=\bfseries,
    ]
    \noindent \textbf{1. From Full-Image to Crop \& Template-First Identification:} \\
    Perform initial reverse image search (Top-N) on the full image. Simultaneously compute scene/frame embeddings and match against built-in \textit{meme-template} and \textit{film-stills} databases (e.g., \textit{The Shining}, \textit{Distracted Boyfriend}, \textit{JoJo `To Be Continued'}).

    $\bullet$ \textbf{Multi-panel/Sequential Narrative Detection}: Detect panel boundaries; extract key actions and causal sequences per panel as distinct evidence nodes.

    $\bullet$ \textbf{Composite/Hybrid Template Detection}: Decompose multi-cue images into components; match each independently and merge candidates with similarity scores.

    \vspace{0.8em}
    \noindent \textbf{2. Prioritize Small-scale UI, Icons, and Symbols:} \\
    Extract high-res crops of small elements (UI icons, logos, symbols, weapons, headwear). Run \textit{icon-lookup} (CN/JP/EN) and record standardized labels---\textbf{no identity assignment}.

    \vspace{0.8em}
    \noindent \textbf{3. Feature Extraction Only:} \\
    For all regions of interest (ROIs), output:
    \begin{itemize}
        \item Bounding box coordinates $(x_1, y_1, x_2, y_2)$
        \item Neutral visual descriptors (e.g., ``red scarf'', ``winged helmet'', ``glowing blue sword'')
        \item On-image text (via OCR, if present)
        \item Non-facial distributive features (weapon shape, emblem, texture, posture)
    \end{itemize}
    \textbf{Do not generate search queries or suggest names.}

    \vspace{0.8em}
    \noindent \textbf{4. Facial \& Sprite-level Comparison:} \\
    If a face/sprite is detected:
    \begin{itemize}
        \item Compute similarity score vs. reference sprites
        \item Require $\ge 2$ independent sources confirming at least one non-facial feature (e.g., emblem + weapon)
    \end{itemize}
    Otherwise: label as \texttt{Tentative} and output ROI bbox for manual review.

    \vspace{0.8em}
    \noindent \textbf{5. Cross-Channel Validation Rule:} \\
    A high-confidence identity assertion requires:
    \begin{itemize}
        \item $\ge 3$ matching distributive features
        \item Verified across $\ge 2$ authoritative sources
    \end{itemize}
    If unmet: retain only bbox + descriptors; \textbf{no name assigned}.

    \vspace{0.8em}
    \noindent \textbf{6. Strict Identity Neutrality:} \\
    \textbf{The system outputs only bounding boxes and objective features---never character/actor names.}
    Identity resolution is deferred to downstream verification (human or external search using the provided bbox + descriptors).
    \end{tcolorbox}
    \caption{Learned protocol for image search.}
    \label{fig:guidance-image-search}
\end{figure*}

\section{Case Study}
\label{app:case-study}

\begin{figure}[t]
    \centering
    \begin{subfigure}[b]{0.48\linewidth}
        \includegraphics[width=\linewidth]{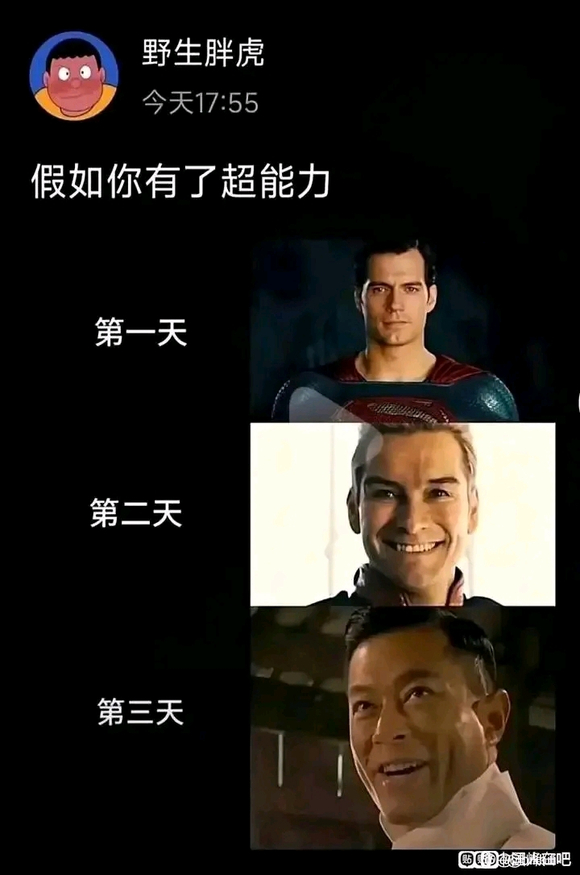}
        \label{fig:meme-left}
    \end{subfigure}
    \hfill
    \begin{subfigure}[b]{0.48\linewidth}
        \includegraphics[width=\linewidth]{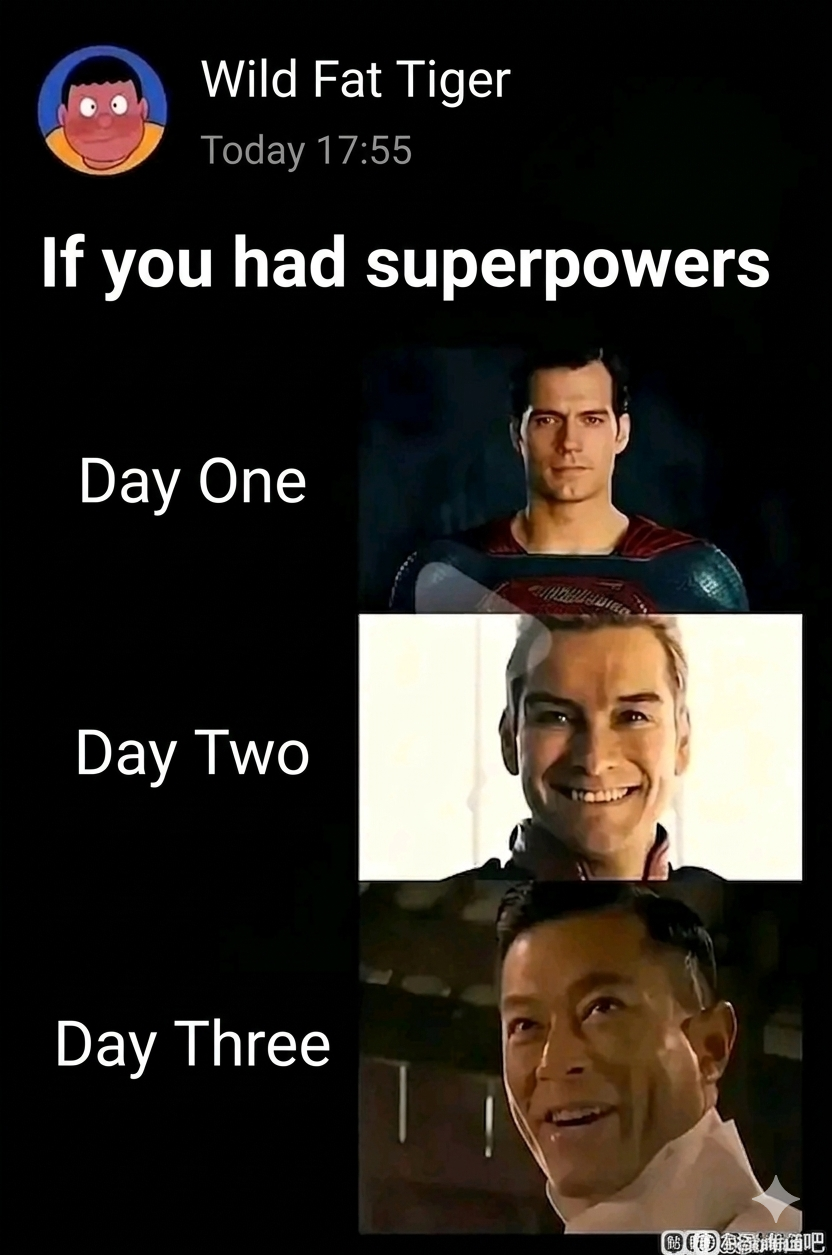}
        \label{fig:meme-right}
    \end{subfigure}
    \caption{Case-study meme: original (left) and translated version (right).}
    \label{fig:meme-example-1}
\end{figure}

This case study examines how the learned guides change evidence acquisition. Figure~\ref{fig:meme-example-1} shows a three-panel Chinese meme combining Superman, Homelander, and Cao Shaolin with the culture-specific phrase ``I don't eat beef.'' A correct explanation requires both visual identity evidence and cross-lingual source context. Figures~\ref{fig:example-ori} and~\ref{fig:example-refined} reproduce the complete executions under the original and learned guides.

Under the original prompt, the agent starts from whole-image facial resemblance and eventually identifies only the generic ``superpower decay'' template. Under the learned guides, it separates the image into panel-level regions, keeps candidate identities provisional, and retrieves independent image evidence for Superman and Homelander. The unresolved phrase becomes the input to a verbatim text search, which links the final panel to Cao Shaolin in \emph{Call of Heroes}.

The resulting evidence supports the final progression from idealized heroic power to authoritarian corruption and then capricious violence. The comparison also isolates the procedural change. The learned context changes the unit of visual search, delays identity commitment until corroborating evidence is available, and passes the unresolved phrase from image retrieval to text search. The traces below preserve every model action and conclusion; only the returned search-result pages are omitted.

\begin{figure*}[htbp]
    \centering
    \begin{adjustbox}{max width=\textwidth,max totalheight=0.86\textheight,center}
    \begin{minipage}{\textwidth}
    \begin{tcolorbox}[
        colback=white,
        colframe=black,
        title=\textbf{Example Result of Origin Prompt},
        fonttitle=\bfseries
    ]
    \textbf{First round of model's output}\\
    \texttt{[reasoning]} The meme shows a Chinese-language post titled (If you had superpowers), followed by three stages: Day 1 (Henry Cavill as Superman, serious), Day 2 (Tom Welling as Clark Kent from \textit{Smallville}, smiling), Day 3 (Andy Lau, grinning mischievously). The progression implies a humorous descent from heroic idealism to playful or self-serving use of powers. The user avatar is (Wild Fat Tiger), referencing the \textit{Doraemon} character Gian (Gian is nicknamed ``Fat Tiger'' in Chinese fan circles), suggesting a humorous, internet-native tone. Key gaps: Who is the third figure (Andy Lau)? Why is he used? Is this a known meme template? Need to verify image identities and cultural context. \texttt{[/reasoning]}\\
    \texttt{[image\_search]}\\
    \{\\
      "rois": [\\
        {}[0.45, 0.27, 0.85, 0.45],\\
        {}[0.45, 0.49, 0.85, 0.68],\\
        {}[0.45, 0.70, 0.85, 0.90]\\
      ]\\
    \}\\
    \texttt{[/image\_search]}

    \bigskip
    \textbf{Second round of model's output}\\
    \texttt{[reasoning]} The image search results confirm the first panel shows Henry Cavill as Superman (from \textit{Man of Steel}), aligning with ``Day 1'' heroic idealism. However, the second and third panels do not match expected identities: the second ROI returns images related to \textit{The Boys} (not Tom Welling/Smallville), and the third ROI links to Chinese internet memes like (``I don't eat beef'') and unrelated figures---not Andy Lau. This suggests either the initial visual identification was incorrect or the meme uses stylized/artistic representations not directly matching real actors. To resolve this, I need to perform a text-based search for the exact Chinese phrase in the meme title and common progression formats to identify the template and cultural context. \texttt{[/reasoning]}\\
    \texttt{[text\_search]}\\
    \{\\
      "queries": ["If you had superpowers meme template", "superpower decay meme Chinese"]\\
    \}\\
    \texttt{[/text\_search]}

    \bigskip
    \textbf{Third round of model's output}\\
    \texttt{[reasoning]} The text search reveals that (``If you had superpowers'') is a known meme template in Chinese internet culture, typically showing a three-stage progression: Day 1 (idealistic/heroic use of powers), Day 2 (more personal or playful use), and Day 3 (absurd, selfish, or humorous misuse). The provided images from the search confirm this structure. However, the specific figures in the user's meme---particularly the third panel---do not match common templates exactly. Given the image search previously failed to identify the third figure as Andy Lau and instead returned unrelated Chinese meme content like (``i don't eat beef'') it is likely the third panel uses a localized, possibly niche or stylized image representing comedic decay rather than a globally recognized celebrity. Since the core meaning of the meme (the humorous decline in superpower usage) is clear from the template structure and Chinese context, and further searches are unlikely to add critical insight within the round limit, I will conclude. \texttt{[/reasoning]}\\
    \texttt{[conclusion]} This meme uses the popular Chinese ``superpower decay'' template titled (``If you had superpowers''), illustrating a humorous three-day progression of how one would use newfound abilities. Day 1 shows Henry Cavill as Superman from \textit{Man of Steel}, symbolizing noble, world-saving intentions. Day 2 transitions to a more relaxed, personal use (though the exact figure is ambiguous, it aligns with a shift toward self-interest). Day 3 features a grinning man---likely a localized internet personality or stock image representing comedic absurdity---signifying the complete abandonment of heroism for trivial or mischievous purposes. Originating from Chinese social media, this template satirizes human nature's tendency to prioritize convenience and fun over responsibility, resonating with global ``power fantasy decay'' memes but tailored to Mandarin-speaking audiences through its language and cultural references. The avatar (Wild Fat Tiger) further anchors it in Chinese netizen culture, referencing the \textit{Doraemon} character Gian to emphasize playful irreverence. \texttt{[/conclusion]}
    \end{tcolorbox}
    \end{minipage}
    \end{adjustbox}
    \caption{Example result of the original prompt. The system prompt and the search results have been omitted. Text enclosed in (``xxx'') appears in Chinese during actual interaction; here it is translated into English.}
    \label{fig:example-ori}
\end{figure*}

\begin{figure*}[htbp]
    \centering
    \begin{adjustbox}{max width=\textwidth,max totalheight=0.88\textheight,center}
    \begin{minipage}{\textwidth}
    \begin{tcolorbox}[
        colback=white,
        colframe=black,
        title=\textbf{Example Result of Prompt with Guidance},
        fonttitle=\bfseries
    ]
    \textbf{First round of model's output}\\
    \texttt{[reasoning]} The full-image search was initiated in the prior step, but no result was returned yet. To proceed efficiently, I must now extract and search individual panels to isolate source material---especially since panel 2 (blond man smiling) and panel 3 (dark-haired man grinning) are likely from distinct film/TV sources that drive the meme's meaning. Per guidance, I should crop each panel and run independent image searches to detect scene embeddings and match against film-still databases.\\
    \\
    I will define three ROIs corresponding to the three panels:\\
    - Panel 1 (Superman): {}[0.40, 0.28, 0.90, 0.48]\\
    - Panel 2 (Blond man): {}[0.40, 0.50, 0.90, 0.70]\\
    - Panel 3 (Dark-haired man): {}[0.40, 0.72, 0.90, 0.98]\\
    \\
    These coordinates capture the central figure in each panel, excluding text labels. Running a multi-ROI image search now enables parallel template detection without assuming identities. \texttt{[/reasoning]}\\
    \texttt{[image\_search]}\\
    \{\\
      "rois": [\\
        {}[0.40, 0.28, 0.90, 0.48],\\
        {}[0.40, 0.50, 0.90, 0.70],\\
        {}[0.40, 0.72, 0.90, 0.98]\\
      ]\\
    \}\\
    \texttt{[/image\_search]}

    \bigskip
    \textbf{Second round of model's output}\\
    \texttt{[reasoning]} The image search results indicate that Panel 1 matches Henry Cavill as Superman from \textit{Man of Steel}, consistent with the ``Day 1'' idealistic hero. Panel 2 is now identified as Homelander from \textit{The Boys}, aligning with a corrupted or authoritarian version of superheroism typical in ``Day 30'' stages. However, Panel 3 only returned associations with the Chinese phrase (``I don't eat beef''), without clear visual or character identification. Since this phrase appears directly in the meme's text, it likely anchors the punchline or cultural reference for Panel 3. To resolve the identity or context of Panel 3, a targeted text search for (``I don't eat beef meme'') is necessary to uncover its origin, usage in memes, and potential link to a specific figure or trope. \texttt{[/reasoning]}\\
    \texttt{[text\_search]}\\
    \{\\
      "queries": [(``I don't eat beef'')]\\
    \}\\
    \texttt{[/text\_search]}

    \bigskip
    \textbf{Third round of model's output}\\
    \texttt{[reasoning]} The text search confirms that (``I don't eat beef'') is a meme originating from the 2016 film \textit{Call of Heroes}, where actor Louis Koo plays the sadistic warlord Cao Shaolin. The character's arbitrary violence---killing people after declaring ``I don't eat beef''---exemplifies capricious tyranny. This aligns with Panel 3's visual: a dark-haired man in military attire grinning maniacally, matching Cao Shaolin's on-screen demeanor. Combined with Panel 1 (Henry Cavill as Superman, representing idealistic heroism) and Panel 2 (Homelander from \textit{The Boys}, symbolizing corrupted superheroism), the meme follows a ``moral decay'' progression: Day 1 = noble hero, Day 30 = authoritarian parody, Day 100 = chaotic, amoral tyrant. The Chinese phrase anchors the final stage in local internet culture, leveraging Louis Koo's iconic role to convey ultimate moral collapse. \texttt{[/reasoning]}\\
    \texttt{[conclusion]} This meme illustrates the progressive corruption of idealism into tyranny through three panels labeled ``Day 1,'' ``Day 30,'' and ``Day 100.'' Panel 1 shows Henry Cavill as Superman from \textit{Man of Steel}, embodying pure heroism. Panel 2 features Homelander from \textit{The Boys}, representing perverted heroism---narcissistic, authoritarian, and violent. Panel 3 depicts Louis Koo as Cao Shaolin from the 2016 Chinese film \textit{Call of Heroes}, whose infamous line (``I don't eat beef'') precedes arbitrary murder, symbolizing complete moral nihilism. The meme uses this Western-to-Chinese cinematic shift to dramatize how power without accountability decays from virtue to madness. The Chinese phrase (``I don't eat beef'') serves as both punchline and cultural shorthand for unpredictable, cruel authority, making the meme resonate particularly in Chinese-speaking online communities. \texttt{[/conclusion]}
    \end{tcolorbox}
    \end{minipage}
    \end{adjustbox}
    \caption{Example result of the prompt with guidance. The system prompt and the search results have been omitted. Text enclosed in (``xxx'') appears in Chinese during actual interaction; here it is translated into English.}
    \label{fig:example-refined}
\end{figure*}